\documentclass[a4paper,fleqn]{cas-dc}

\usepackage[numbers]{natbib}
\usepackage{pifont}

\def\tsc#1{\csdef{#1}{\textsc{\lowercase{#1}}\xspace}}
\tsc{WGM}
\tsc{QE}

\begin{document}
\let\WriteBookmarks\relax
\def\floatpagepagefraction{1}
\def\textpagefraction{.001}
\shorttitle{Ultrasound  Video Segmentation via MWNet}
\shortauthors{C, Zhang et~al.}

\title [mode = title]{Tracking spatial temporal details in ultrasound long video via wavelet analysis and memory bank}                   
\tnotemark[1]

\tnotetext[1]{This work was supported in part by the National natural Science Foundation of China under Grant 62173014 and Grant U22A2051, in part by the National Key Research and Development Program of China under Grant 2022YFC2405401, and in part by the Natural Science Foundation of Beijing Municipality under Grant L232037 and L242166. (Chenxiao Zhang and Runshi Zhang contributed equally to this work.) (Corresponding author: Junchen Wang.)}


\author[]{Chenxiao Zhang$^1$}

\author[]{Runshi Zhang$^1$}

\author[]{Junchen Wang\corref{cor1}}[
orcid=0000-0002-0916-9932]

\affiliation[]{organization={School of Mechanical Engineering and Automation, Beihang University},
	addressline={37 Xueyuan Road, Haidian District}, 
	postcode={100191}, 
	state={Beijing},
	country={China}}

\cortext[cor1]{Corresponding author}
\fntext[fn1]{Co-first authors.}
\ead{wangjunchen@buaa.edu.cn}

\ead[url]{https://mrs.buaa.edu.cn/}



\begin{abstract}
Medical ultrasound videos are widely used for medical inspections, disease diagnosis and surgical planning. 
High-fidelity lesion area and target organ segmentation constitutes a key component of the computer-assisted surgery workflow.
The low contrast levels and noisy backgrounds of ultrasound videos cause missegmentation of organ boundary, which may lead to small object losses and increase boundary segmentation errors.
Object tracking in long videos also remains a significant research challenge.
To overcome these challenges, we propose a memory bank-based wavelet filtering and fusion network, which adopts an encoder-decoder structure to effectively extract fine-grained detailed spatial features and integrate high-frequency (HF) information.
Specifically, memory-based wavelet convolution is presented to simultaneously capture category, detailed information and utilize adjacent information in the encoder. 
Cascaded wavelet compression is used to fuse multiscale frequency-domain features and expand the receptive field within each convolutional layer.
A long short-term memory bank using cross-attention and memory compression mechanisms is designed to track objects in long video.
To fully utilize the boundary-sensitive HF details of feature maps, an HF-aware feature fusion module is designed via adaptive wavelet filters in the decoder. 
In extensive benchmark tests conducted on four ultrasound video datasets (two thyroid nodule, the thyroid gland, the heart datasets) compared with the state-of-the-art methods, our method demonstrates marked improvements in segmentation metrics.
In particular, our method can more accurately segment small thyroid nodules, demonstrating its effectiveness for cases involving small ultrasound objects in long video.
The code is available at \underline{https://github.com/XiAooZ/MWNet}.
\end{abstract}




\begin{highlights}
\item We proposed a memory bank- and wavelet-based network for segmentation.
\item It can emphasize the fine-grained details and small object in long ultrasound video.
\item The memory-based wavelet convolution was presented to fuse spatiotemporal feature.
\item The long short-term memory bank is designed for modeling long-term dependencies.
\item We proposed the high frequency-aware feature fusion to fuse multi-scale feature.
\end{highlights}


\maketitle
\section{Introduction}
\begin{figure}[!t]
	\centerline{\includegraphics[scale=0.25]{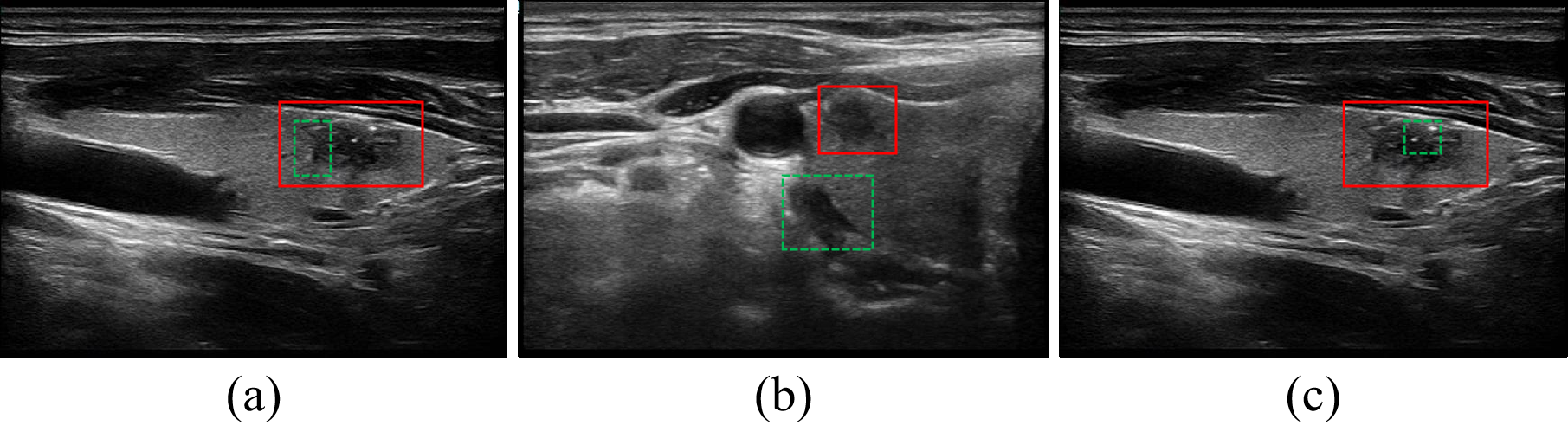}}
	\caption{The challenges of ultrasound video segmentation: (a) blurred boundaries, (b) confusing locations, (c) speckle noise, red contour refers to the thyroid nodule and green contour refers to the confusing one.}
	\label{fig1}
\end{figure}

Medical ultrasound videos have been extensively used for the screening of diseases such as thyroid nodules~\cite{ThyroidNodule1}, breast lesions~\cite{BreastLesion}, and heart disease~\cite{MemSAM}, etc. 
When doing so, precisely segmenting the target tissues and diseased areas is essential for diagnosis, surgery planning, and intraoperative and postoperative observations~\cite{LayerSupervise}.
To this end, manual labeling for the diseased area is often employed in conventional approaches, which is time consuming and susceptible to mistakes.
Therefore, automatic ultrasound video segmentation is highly significant for diagnostic efficiency and reducing burden imposed on surgeons. 
Ultrasound image segmentation often faces challenges arising from the inherent image characteristics, such as low contrast, blurred boundaries, confusing locations, and speckle noise~\cite{SegmentationChallenge}, as illustrated in Fig~.\ref{fig1}. 
Among them, low contrast levels and blurred boundaries cause boundary displacement, whereas confusing locations may lead to missed segmentation.

Recently, deep neural networks have demonstrated significant success in automatic medical segmentation tasks.
U-Net, a classic segmentation framework, is widely used in medical image because of its simple but effective design.
Subsequent studies~\cite{AAU-Net, unet++} have focused on individually optimizing the encoder and decoder of U-Net to further improve its segmentation accuracy.
We follow this strategy and further incorporate temporal information modeling. 
It has been demonstrated that fine-grained spatial details play a critical role in precise boundary segmentation~\cite{finegrainedaccuracy}, particularly for low quality ultrasound videos.
Furthermore, recent studies~\cite{FreqFusion} indicate that such details, mainly represented in shallow feature maps, correspond to high-frequency (HF) components within the frequency domain.
Convolutional neural networks (ConvNets) offer superior performance in preserving these HF components ~\cite{ConvHigh1}, while transformer-based methods have been proposed to more effectively capture the long-range dependencies among medical images~\cite{zhang2024utsrmorph, zhang2025tcfnet}.
Therefore, the current method~\cite{CNN+Transformer2} simply integrates the Swin Transformer (Swin-T)~\cite{SwinTransformer} and ResNet backbones to combine global context information with local details. 
Nevertheless, this strategy leads to information redundancy and the insufficient utilization of HF boundary features. 
In summary, developing a robust backbone that can extract both long-range spatial information and fine-grained local features is essential to enhance segmentation performance.
Furthermore, several frequency-domain methods have been developed to enhance the utilizing of multiscale information and the extracting of HF features~\cite{chen2025frequency}.
The wavelet transform, which was designed for multiresolution and time-frequency analyses~\cite{WTConv}, is particularly suitable for local detail preservation and global context modeling.
Cascaded wavelet decomposition can progressively enlarge the receptive field through feature decomposition and recombination, thereby enabling the rich global contextual representations.
To address the limitations of ConvNets in capturing fine-grained HF information, we incorporate the wavelet transform into our method. 
This integration allows for enriched HF feature modeling while simultaneously promoting global contextual awareness during both feature extraction and fusion.

Earlier approaches adopted online learning paradigms to exploit neighborhood memory features for modeling temporal dependencies~\cite{FLANet}~\cite{DPSTT}.
It is evident that short-term memory mechanisms are less effective for long video object tracking, particularly when dealing with small object.
Recent approaches~\cite{MemSAM}~\cite{Vivim} primarily rely on memory banks and pixel-level attention mechanisms for memory reading~\cite{zhou2024rmem}.
Inspired by them, we use convolutional long short-term memory (ConvLSTM)~\cite{ConvLSTM} to capture multiscale short-term features in adjacent frames, and a memory bank with pixel-level memorization and updating is proposed to preserve long video information.
They can enable HF spatiotemporal modeling and significantly improve long-range dependencies.

In this paper, we propose a memory bank-based wavelet filtering and fusion network (MWNet) to precisely segment lesion regions and target tissues in ultrasound long videos, which adopts an encoder-decoder structure. 
Specifically, the main contributions of this work can be summarized as follows:

(1) 
 A memory-based wavelet convolution (MWConv) backbone is developed to extract multiscale spatial and temporal features from the wavelet domain. 
 	It employs cascaded wavelet compression to fuse the multiscale frequency-domain features within each convolutional layer and ConvLSTM to enhance HF information in multiscale feature maps.

(2) A long short-term memory bank using cross-attention and memory compression mechanisms is proposed to combine the temporal correlations between both nearby and distant frames and the current frame, then facilitating long-term tracking performance.

(3) An HF-aware feature fusion (HFF) module that adopts adaptive wavelet filters, is proposed to concentrate on the different frequency components contained in multiscale feature maps, suppress noise and highlight the HF components. 

According to experimental results obtained in different medical video segmentation tasks (four datasets), including those involving thyroid nodules, the thyroid gland, and the heart, our method achieves significant improvements on multiple evaluation metrics compared with other state-of-the-art (SOTA) ultrasound segmentation approaches.
\begin{figure*}[!t]
	\centerline{\includegraphics[width=\textwidth]{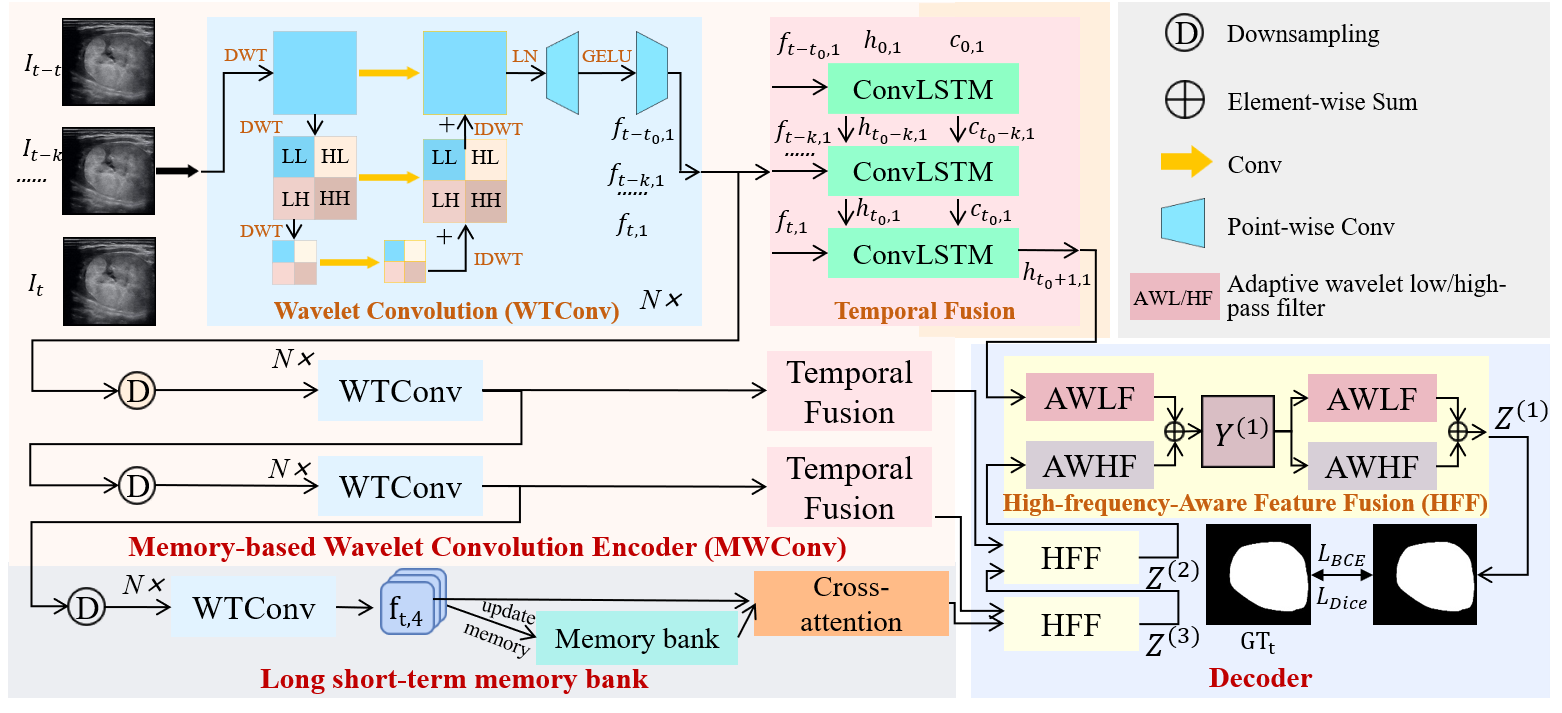}}
	\caption{The MWNet includes three stages in the encoder, a memory bank and several HFF modules in the decoder. Each stage consists of several shared wavelet convolution block (WTConv) and a temporal fusion module. In WTConv, the input feature maps are first decomposed through a cascaded wavelet transform to achieve multi-resolution representations. The feature reconstruction is performed using the IWT. The temporal fusion module leverages ConvLSTM to achieve temporal feature integration of adjacent frames. The long short-term memory bank is used to track segmentation object for the deepest feature maps. In HFF, each module takes a low and a high resolution feature map as input, with the fused output in high resolution. }
	\label{fig2}
\end{figure*}

\section{Related Works}

\subsection{Ultrasound Segmentation}
Deep learning has driven remarkable advancements in ultrasound video segmentation~\cite{DeeplearningUltraosound}. 
Numerous methods have been proposed to address the segmentation challenges caused by blurred boundaries and unclear tissues. 
Mishra et al.~\cite{LayerSupervise} proposed a fully supervised network that computes the boundary and object losses between object regions and the background in its multiresolution feature layers, improving the boundary segmentation accuracy.
Chen et al.~\cite{AAU-Net} replaced the traditional convolutional operation with a hybrid adaptive attention module, including channel and spatial attention blocks, to capture more features with different receptive fields.
Ning et al.~\cite{EdgeBackgroundAware} devised a morphology-aware framework that improves the ability of the constructed model to learn detailed spatial information such as boundaries. 
Wang et al.~\cite{boundaryloss} combined ConvNet, multilayer perceptron (MLP) and pyramid structures to extract global contextual information and improve segmentation accuracy.
With the widespread application of deep learning in video understanding scenarios, ultrasound video segmentation has evolved from a static image processing problem to sophisticated spatiotemporal analysis of ultrasound video.
Lin et al.~\cite{FLANet} effectively exploited the frequency information contained in adjacent frames by transferring their feature maps to the frequency domain and then performing a temporal fusion on them. 
Li et al.~\cite{DPSTT} improved the temporal tracking process by integrating a memory bank with decoupled transformers, enabling lesion movements to be efficiently tracked in ultrasound video over time. 
Although the mainstream ultrasound video segmentation methods primarily focus on temporal information, other approaches are aimed at developing feature extraction and fusion modules based on ultrasound image properties to achieve accurate segmentation.
Hu et al.~\cite{RMFG_net} proposed a feature filter and a gated fusion module to suppress noise and enhance segmentation performance of the detailed area.
These studies indicate that focusing on boundary and utilizing multiscale spatiotemporal information are crucial for segmenting ultrasound videos.

\subsection{Wavelet Transform in Segmentation Tasks}
As a powerful tool for multiresolution and time-frequency analyses, the wavelet transform has been integrated into deep learning methods to enable the comprehension of detailed spatial information at multiple scales~\cite{DAWN}.
Xu et al.~\cite{WaveletDownsample} implemented a learnable downsampling method leveraging the Haar wavelet, which can reduce the spatial resolution and retain more information during the downsampling process. 
Wen et al.~\cite{GaborWavelet} incorporated a Gabor wavelet into ConvNets to extract texture information at different scales and directions. 
Bi et al.~\cite{FoggyWavelet} proposed a bidirectional wavelet guidance mechanism to simultaneously enhance low-frequency (LF) and HF features. 
Zhou et al.~\cite{Gobletnet} presented a parallel HF encoder to effectively extract detail information, using the wavelet transform to generate HF features as extra inputs. 
Overall, the wavelet transform enables the effective extraction of multiscale information, then facilitating the selective utilization of frequency-domain feature components on the basis of specific segmentation requirements.

For ultrasound image/video segmentation task, 
D. Li~\cite{dandan2023semi} mechanically integrates the wavelet decomposition and inverse wavelet transform (IWT) into the stages in the encoder and decoder of UNet, respectively. 
Y. Xiong~\cite{xiong2025hcmnet} implemented the wavelet decomposition in parallel with the convolutional encoder and combined features from both modules at different resolutions through the Mamba module. 
S. Zheng~\cite{zheng2024gwunet} incorporated the wavelet decomposition module into the skip connection of the shallowest layer to capture precise nodule regions. 
P. Liang~\cite{liang2026wkpnet} incorporated only a single wavelet decomposition module into existing networks (Kolmogorov-Arnold Network and Polarized Linear Attention) to enhance multi-resolution feature extraction. 
Recently, X. Luo~\cite{luo2025lgffm} proposed a wavelet decomposition-based frequency domain mapping module after encoder to extract HF and LF feature maps. 
Overall, most existing approaches integrate wavelet modules as plug-and-play components within a ConvNet-based or Transformer-based encoder-decoder framework, commonly including skip connections~\cite{zheng2024gwunet} (particularly at the deepest layers~\cite{luo2025lgffm}~\cite{liang2026wkpnet}), parallel pathways~\cite{xiong2025hcmnet} or concatenation strategies~\cite{dandan2023semi} in the encoder. 
However, existing methods often treat wavelet module as a plug-and-play module and underestimate the critical importance of the feature fusion stage in the decoder for accurately capturing fine-grained segmentation boundaries. 
In contrast, we propose a purely wavelet-based design that incorporates dedicated wavelet-based layers in both the encoder and decoder. 
We combine the wavelet decomposition and IWT operations within the encoder processing pipeline and employ wavelet-based multifrequency feature fusion with special attention given to HF information. 
Specifically, fixed-basis wavelets in the encoder can expand the receptive field and generate multi-scale feature through cascaded LF decomposition. The adaptive wavelets in the decoder can achieve adaptive low/high-pass filter to suppress noise and highlight HF details.

\subsection{Memory-based Video Understanding}
The memory-based module~\cite{zhou2024rmem} constitutes a central component in ultrasound object segmentation, offering a robust mechanism to alleviate object loss in long video sequences while effectively controlling redundant information. 
Mainstream memory-based modules~\cite{MemSAM} include adjacent frame aggregation and memory bank. 
The former can integrate information from neighboring frames, whereas the latter incrementally incorporates sampled frames throughout the video processing pipeline. 
Early studies~\cite{FLANet,lin2024instrument,DPSTT} predominantly focused on adjacent frame aggregation, employing models such as LSTMs to capture the spatiotemporal dependencies between the current frame and its preceding frames. 
X Deng~\cite{deng2025echocardiography} focused on neighborhood correlation of adjacent two frames in echocardiography video. 
Consequently, memory bank-based methods have attracted growing interest owing to their advantages in retaining salient information from previous frames. 
Z. Yang~\cite{aot} adopted a transformer-inspired architecture to build a memory bank, incorporating long short-term attention modules after the self-attention operation to capture relationships among current frame features, long short-term historical features, and segmentation masks, thereby facilitating multi-object segmentation. 
However, its long-term attention mechanism mechanically accumulates information from all frames, thereby restricting scalability with respect to the sequence length. 
J. Zhou~\cite{zhou2024rmem} proposed restricted memory banks with an update mechanism to handle redundant information. 

For ultrasound segmentation task,
H. Zhao~\cite{zhao2023ultrasound} proposed confidence-based memory update and the confidence score computed from skip gate module, but its performance depends on keyframe annotations. 
J. Li~\cite{li2024cascaded} partitioned long videos into multiple clips for the network input and designed two Transformer-based modules after encoding to capture spatial-temporal information within each clip and across the memory clip, thereby enhancing tumor localization accuracy. 
However, the deepest layer of the encoder retains only limited fine-grained positional information.
X. Deng~\cite{MemSAM} proposed MemSAM for echocardiography video segmentation, which employed a space-time memory bank as prompt for the current frame and presented a memory update mechanism to address speckle noise. 
However, it relies on limited annotations and the memory banks are often deficient in preserving HF feature representations.
Y. Yang~\cite{Vivim} adopted state space models (SSMs) to model long-term spatiotemporal features across multiple feature scales, which achieved excellent performance in several ultrasound videos. 
However, the simplistic accumulation of long-term memory features leads to a rapid increase in memory consumption with longer sequences, which consequently diminishes the method’s effectiveness in processing extended videos.
Z. Tu~\cite{tu2025spatial} proposed a novel spatial-temporal memory filtering SAM network and its dense prompts are from three relevant reference frames (the first frame with the ground truth mask, the most recent frame and the previous frame with the highest similarity). 
In contrast, this study proposed a dynamically updated long short-term memory bank to model long-range temporal dependencies without the annotation supervision in ultrasound long video segmentation.
\subsection{Feature Fusion in Segmentation}
Feature fusion plays a pivotal role in semantic segmentation by effectively combining low-resolution feature maps with high-resolution semantic information. 
The refinement of upsampling operations represents a key direction for enhancing the feature fusion, for example interpolation upsampling using fixed/hand-crafted kernels, deconvolution~\cite{zeiler2014visualizing} and pixel-shuffle~\cite{PixelShuffle} using learnable kernel parameters. 
Subsequently, some researchers focused on dynamic upsampling operators~\cite{CARAFE,SAPA,FADE} to achieve better feature upsampling. 
DySample~\cite{Dysample} adopted sampling-based design to learn offset and align the low and high-resolution features. 
Recently, FreqFusion~\cite{FreqFusion} demonstrated frequency component analysis using conventional convolution theorem can improve multi-scale feature fusion and segmentation accuracy, especially for HF. 
WeaveSeg~\cite{li2025weaveseg} employed adaptive spectral refinement within the feature fusion module to enhance HF boundary details, producing nuclei that are both semantically coherent and sharply delineated for nuclei instance segmentation in histopathology images. 
Inspired by them, the integration of wavelet transform theory facilitates adaptive filtering and upsampling fusion in this paper, leading to improved HF boundaries and enhanced suppression of background noise.
\section{Methodology}
\subsection{Overview}
In this study, we propose a novel end-to-end ultrasound video segmentation framework, termed MWNet, which is shown as illustrated in Fig.~\ref{fig2}. 
The model is built upon an encoder-decoder structure and comprises three key components:
a memory-based wavelet convolution (MWConv) encoder, a long short-term memory bank and HF-aware feature fusion (HFF) decoder.
Specifically, a shared hierarchical wavelet convolution (WTConv) backbone is first employed to extract multiresolution feature maps of the current and previous frames from an ultrasound video.
A temporal fusion module is subsequently introduced to capture the temporal information between the adjacent frames.
These two operations together constitute the MWConv encoder.
Following the stage, the feature maps corresponding to the current frame are selected for further refinement. 
After the deepest stage, a long short-term memory bank is proposed to aggregate long-, short- range temporal information.
In decoder, the features are passed through a series of HFF modules, which are tailored to enhance structural details by emphasizing HF components.
\subsection{Memory-based Wavelet Convolution Encoder}
\label{Encoder}
The ConvNet-based method encounters two fundamental challenges: 
an insufficient ability to capture HF information across multiple scales and resolutions;
its dependence on large convolutional kernels for effectively enlarging receptive fields, leading to increased GPU memory consumption levels. 
To address these challenges simultaneously, wavelet transform techniques are incorporated into our backbone to exploit their sensitivity to HF components for multiscale feature extraction.
And cascaded wavelet decomposition is utilized to effectively perform cross-resolution feature fusion, ultimately increasing both receptive field size and feature richness.
The proposed backbone is designed to be applicable to both ultrasound images and videos.
To effectively capture the temporal dependencies between adjacent frames in ultrasound videos, a temporal fusion module is incorporated into the framework.
By jointly modeling spatial and temporal features, a unified backbone is established to handle ultrasound videos.

\subsubsection{Overall Structure}
It extends the ConvNeXt~\cite{ConvNeXt} by embedding WTConv for frequency-aware spatial modeling and introducing temporal fusion modules to capture interframe dependencies and it is as shown in Fig.~\ref{fig2}. 
The MWConv encoder adopts a hierarchical design composed of three stages. 

\subsubsection{Wavelet Convolution}
A wavelet transform operation is used to decompose the feature maps ${X}^{(i)}$, then generating,
\begin{align}
	{X}^{(l,1)}_{LL}, {X}^{(l,1)}_{LH}, {X}^{(l,1)}_{HL}, {X}^{(l,1)}_{HH} &= {\text{WT}({X}^{(l)})}
\end{align}
where the four components represent the LF approximation (LL), horizontal detail (LH), vertical detail (HL), and diagonal detail (HH), respectively. 
WT is the wavelet transform operation, $l$ is the $l$-th stage in encoder, and $l=1,2,3,4$. 
The last three of these components can be combined and simplified as ${X}_{H}^{(l,1)}$.
To further enhance multiscale representation in the $l$-th stage, the LF features ${X}^{l,1}_{LL}$ are recursively decomposed to obtain multiresolution feature maps with several wavelet transform operations, which continue until the lowest resolution equals 1/4 of the deepest resolution of the encoder.
\begin{align}
	{X}^{(l,k+1)}_{LL}, {X}^{(l,k+1)}_{H} &= {\text{WT}({X}^{(l,k)}_{LL})}
\end{align}
where $k=1, 2, ..., 6-l$, indicates the number of recursive wavelet decomposition.
Then, small kernel convolutions are applied to the downsampling feature maps to obtain feature maps ${X}^{(l,k+1)}_{LL^1},{X}^{(l,k+1)}_{H^1}$, which can achieve an enlarged effective receptive field and offer a more memory-efficient alternative to large-kernel convolutions.
The lowest LL resolution feature maps are ${X}^{(l,7-l)}_{LL^1}$, which is defined as ${X}^{(l,7-l)}_{LL^2}$.

IWT is subsequently employed to progressively upsample the low-resolution features, thereby fusing and generating higher-resolutions.
Specifically, at each scale, hierarchical merging is performed by adding the IWT resulting from the lower resolution and the LF component derived from the higher resolution in an elementwise manner. 
It can be formally written as follows:
\begin{align}
	{X}_{LL^2}^{(l,k-1)} = \text{IWT}({X}^{(l,k)}_{LL^2},{X}_{H^1}^{(l,k)})+{X}^{(l,k-1)}_{LL^1}
\end{align}
where $k=7-l, 6-l, ..., 2$, IWT is inverse wavelet transform operation.
The output of the block is,
\begin{align}
	{X}_{1}^{(l)} = \text{IWT}({X}^{(l,1)}_{LL^2},{X}_{H^1}^{(l,1)})+\text{Conv}({X}^{(l)})
\end{align}
Conv here is a depthwise convolution layer.

\subsubsection{Temporal Fusion}
The temporal fusion module takes the feature maps derived from several consecutive frames at the current level as its input to model temporal dependencies, and it is as shown in Fig.~\ref{fig2}.
ConvLSTM is chosen as the core component.
Taking stage $l$ as an example, temporal correlations are established by modeling the feature relationships between the current frame $f_{t,l}$ and its preceding frames $f_{t-1,l}, f_{t-2,l}, ..., f_{t-t_0,l}$,
\begin{equation}{h}_{t_0-k+1,l}, {c}_{t_0-k+1,l} = \text{ConvLSTM}({f}_{t-k,l},{h}_{t_0-k,l},{c}_{t_0-k,l})\label{eq20}\end{equation}
where $k=t_0, t_0-1, ..., 0$.
Each ConvLSTM cell receives the cell state ${c}_{t_0-k,l}$ and the hidden state ${h}_{t_0-k,l}$ as temporal inputs, along with the feature map of the corresponding frame at the current hierarchical level ${f}_{t-k,l}$ as the current input. 
The cell models temporal dynamics of these inputs and updates the internal states, producing a new cell state ${c}_{t_0-k+1,l}$ and a hidden state ${h}_{t_0-k+1,l}$. 
The final cell state ${h}_{t_0+1,l}$ represents the fused temporal features of the corresponding layer across the previous frames and serves as the output of the module at that layer, which is denoted as ${X}_{l}$.
\begin{figure}[!t]
	\centerline{\includegraphics[width=\columnwidth]{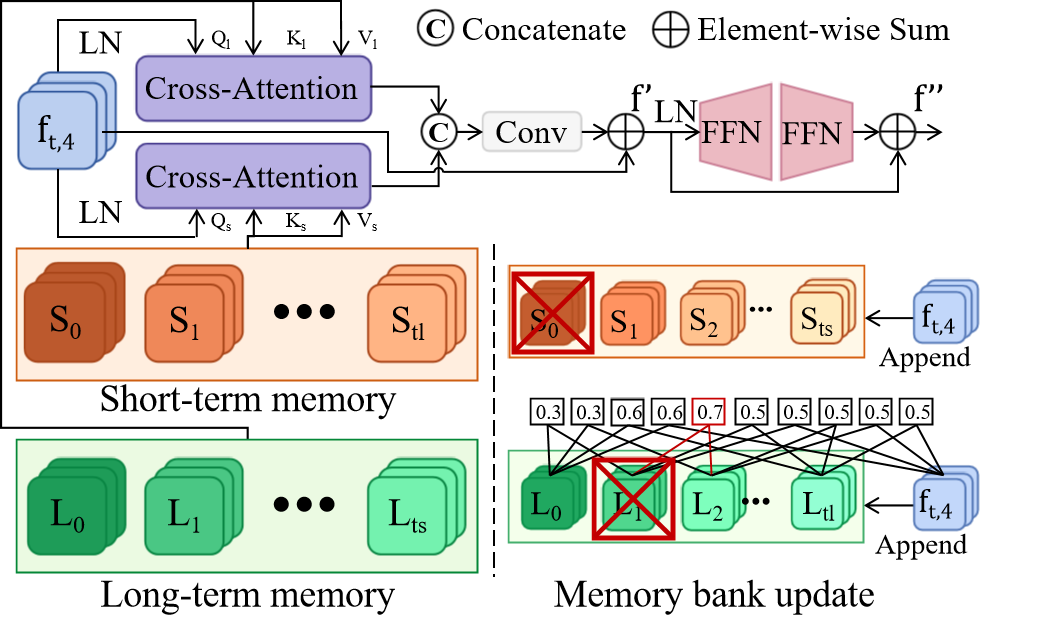}}
	\caption{The detail of long short-term memory bank.}
	\label{memory}
\end{figure}
\begin{figure*}[!t]
	\centerline{\includegraphics[width=\textwidth]{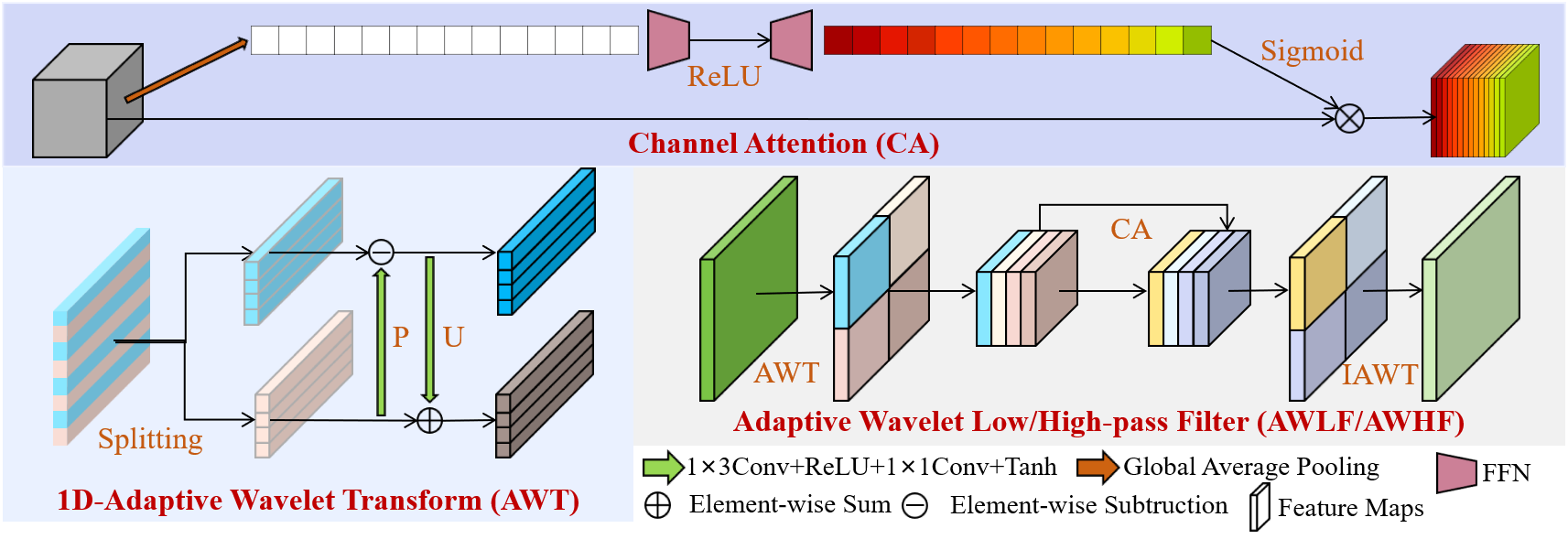}}
	\caption{The details of adaptive wavelet low/high-pass filter (AWLF/AWHF). The feature map is first decomposed by a two-dimensional adaptive wavelet transform (AWT). Specifically, the 2D-AWT is realized by executing horizontal and vertical 1D-AWT successively. 
		Subsequently, the wavelet decompositions are reweighted using a channel attention module, and the reconstruction is ultimately performed by a 2D inverse adaptive wavelet transform (2D-IAWT), which serves as the inverse operation of the forward 2D-AWT.}
	\label{fig3}
\end{figure*}
\subsection{Long short-term memory bank}
The memory banks are essential for establishing the relationship between the current input and past features, particularly when processing long videos, and it serves as the foundation for video object segmentation~\cite{he2024ma, zhou2024rmem, yang2021associating}.
It integrates past object segmentation features and employs an update mechanism to adapt to long video sequences.
In MWNet, as shown in Fig.~\ref{memory}, a long short-term memory bank is designed for modeling the long-range dependency and short-term variance between the current input and past features simultaneously.
In detail, the input current feature ${f}_{t,4}$ is initially computed via cross-attention mechanism with the long- and short-term memory banks, respectively, to capture their mutual relationship.
Take the long-term cross-attention as an example, each feature in memory bank is first concatenated on the channel dimension:
${m}_{l}= \text{Cat}({L}_{0}, {L}_{1}...{L}_{l})$.
Subsequently, the query, key and value are computed via three weight matrix, where the query is calculated based on the current feature, and the key and value are obtained from the memory $m_l$.
Then the cross-attention is applied as:
\begin{align}
	{f}_{l} = \text{Attn}({Q}_{t}, {K}_{l}, {V}_{l}) = \text{Softmax}(\frac{Q_{t}K_{l}^{T}}{\sqrt{C}})V_{l}
\end{align}
Where LN is a layernorm operation, $f_l$ is the long-term dependencies features, $C$ is the number of channels. Likewise, the feature map $f_s$ can be derived by substituting long-term memory $m_l$ with short-distance memory $m_s$ in the above process.
A pointwise convolution is next applied to the concatenated outputs to achieve channel alignment.
Finally, the representation capability is enhanced via a feedforward network.

To control GPU memory consumption, we restrict the length of both long- and short-term memory banks, and they are updated using different strategies.
The restricted lengths are $t_l$ and $t_s$, respectively. 
The features are progressively compressed and stored in the memory bank in the initial frame.
If the short-term memory bank exceeds its length constraint, the earliest features are removed and the current features are appended. 
To ensure that early-stage information is retained and the long-term memory captures a holistic representation of the video, we introduce a similarity-driven memory update mechanism, which is shown in Fig.~\ref{memory}. 
Upon appending the current features into the memory bank, pairwise feature similarities are calculated, and the high redundancies are selectively discarded.

\subsection{HF-aware feature fusion}
\label{Fusion}
Filtering and fusing multiscale feature maps enriched with multifrequency information are beneficial for attaining improved segmentation accuracy. 
Previous work~\cite{highfrequency2intra} has demonstrated that disrupting HF components leads to a notable drop in intracategory similarity, resulting in degraded feature representations. 
Moreover, ultrasound video segmentation, especially in cases with blurred boundaries and small objects, depends on leveraging fine-grained details, which typically appear as an HF component in the feature map.
Shallow feature maps have been proven to contain more fine-grained details, whereas deep feature maps capture richer deep semantic information~\cite{FreqFusion}.
Therefore, we introduce a HF-aware feature fusion (HFF) module based on the wavelet transform to maintain the essential details while filter out noisy HF components,
then utilizing the multiscale features of the encoder and enhancing the HF information.

\subsubsection{Overall Structure}
A previous study~\cite{FreqFusion} establishes that enhancing the initial fusion of HF components plays a crucial role in enabling the adaptive feature filter to achieve better performance.
Building on this insight, the adaptive wavelet-based low and high-pass filter (AWLF and AWHF) are incorporated to refine the feature fusion outputs of the previous layer ${Z}^{l+1} \in \mathbb{R}^{{C}\times \frac{H}{2} \times \frac{W}{2}}$ and the residual connection derived from the corresponding encoder layer ${X}^{l}\in \mathbb{R}^{{C}_{l}\times H \times W}$.
The filtered results ${Y}^{l}\in \mathbb{R}^{{C}\times H \times W}$ are subsequently combined through a simple yet effective elementwise addition operation, yielding the initial filtered features. 
Pointwise convolutions are adopted for feature alignment and channel compression.
\begin{align}
	{Y}^{(l)} = \mathcal{F^\mathrm{L}}(\mathcal{F^\mathrm{UP}}(\text{Conv}({Z}^{(l+1)}))) + \mathcal{F^\mathrm{H}}({\text{Conv}}({X}^{(l)}))
\end{align}
where $\mathcal{F^\mathrm{L}}$ and $\mathcal{F^\mathrm{H}}$ indicate the AWLF and AWHF respectively, $\mathcal{F^\mathrm{UP}}$ denotes the bilinear interpolation upsampling operation, and Conv here indicates the pointwise convolution utilized for channel alignment.

To achieve more precise HF perception effects, through fine-grained feature filtering, we subsequently utilize a second pair of AWLF/AWHF to remodel the frequency domain of the initial fusion outputs. The resulting features are subsequently aligned along the feature dimension via pointwise convolution.
\begin{align}
	{Z}_{l} = \text{Conv}(\mathcal{F^\mathrm{L}}({Y}^{(l)})) + \text{Conv}(\mathcal{F^\mathrm{H}}({Y}^{(l)}))
\end{align}
where ${Y}^{(l)}$ is the result of the initial fusion phase.
Leveraging the prior knowledge that shallow feature maps are rich in HF information, as shown in Fig.~\ref{fig3}, we utilize a high-pass filter to preserve the discriminative fine-grained features, while a low-pass filter effectively suppresses disruptive noise, leading to a HFF. 
The details of the proposed modules are introduced below.

\subsubsection{Adaptive Wavelet Transform}
The implementation of the traditional wavelet transform relies on a fixed wavelet basis. The signal's HF and LF components are extracted by convolving it with the wavelet and scaling functions, respectively.
Therefore, this process relies on a manually selected wavelet basis, which is unable to fully accommodate the diverse characteristics of the feature maps at different layers.
The lifting scheme~\cite{LiftingScheme}, also known as the second-generation wavelet transform, provides a novel approach for implementing wavelet transforms and serves as a foundation for the development of a data-driven adaptive wavelet transform (AWT). 
The original version of lifting scheme is nonadaptive, since the predictor function P and updater functions U, which are crucial to the transform mechanism of the lifting scheme, are fixed in advance.
To address this limitation, these two functions are instead parameterized and made learnable, thereby enabling an AWT.
In general, the predictor and updater functions rely on localized operations implemented over the decomposition results, for example by computing the mean or fitting a model over a small neighborhood of discrete points. 
Building on this observation, the AWT is implemented by replacing the fixed functions P and U with two one-dimensional convolution operations. 

A discrete one-dimensional signal $x$ is first split according to the parity of its elements. 
\begin{align}
	x_{e}[n] = x[2n],\quad x_{o}[n] = x[2n+1]
\end{align}
where $x_{e}$ and $x_o$ indicate the even- and odd- indexed components, respectively.
A predictor function is subsequently utilized to approximate $x_{o}$ on the basis of $x_{e}$. The resulting prediction is subtracted from the original odd components to extract the detail coefficients, which correspond to the HF content of the signal $H$. 
\begin{align}
	H = x_{o} - \text{{Conv}}_\text{P}(x_{e})
\end{align}
where $\text{Conv}_\text{P}$ denotes the predictor function which is a convolution operation in this study.
Finally, an updater function is employed to modify the even-indexed components using the previously computed HF coefficients. 
Through this update operation, a low-pass approximation $L$ is obtained.
\begin{align}
	L = x_{e} + \text{Conv}_\text{U}(H)
\end{align}
where $\text{Conv}_\text{U}$ denotes the updater function which is a convolution operation in this study.
The inverse adaptive wavelet (IAWT) is first realized by applying the principles of the lifting scheme.
\begin{align}
	x_{e} = L - \text{Conv}_\text{U}(H),\quad
	x_{o} = H + \text{Conv}_\text{P}(x_{e})
\end{align}
Furthermore, the AWT is extended to two dimensions by performing it successively in both the horizontal and vertical directions. 

\subsubsection{Adaptive Wavelet Filter}
The feature map used to filter $Y$ is first decomposed into four subbands through the AWT:
\begin{align}
	{Y}_{LL}, {Y}_{LH}, {Y}_{HL}, {Y}_{HH}=\text{AWT}({Y})
\end{align}
where AWT  denotes the two-dimensional AWT operation, and ${Y}_{LL}, {Y}_{LH}, {Y}_{HL}, {Y}_{HH}$ indicates the multifrequency decomposition components obtained by the AWT.
We incorporate a channel attention (CA) module, which is a squeeze and excitation (SE) block~\cite{SENet} in MWNet, to model the channel relationships within each wavelet decomposition component and perform a reweighting operation. 
Taking high-pass filtering as an example, higher weights are assigned to the HF subbands, whereas in low-pass filtering, the LF subbands are prioritized. 
\begin{align}
	{Y}_{LL}^{F}, {Y}_{LH}^{F}, {Y}_{HL}^{F}, {Y}_{HH}^{F}=\text{CA}({Y}_{LL}, {Y}_{LH}, {Y}_{HL}, {Y}_{HH})
\end{align}
where CA indicates the channel attention performed obtained by the SE block, the ${Y}_{LL}^{F}$ refers to the weighted LF decomposition, and the remaining terms in the form of ${Y}^F$ correspond to the other weighted decomposition components. 
The weighted subbands are finally reconstructed using an IAWT to obtain the final filtered result.
\begin{align}
	{Y}^{F}=\text{IAWT}( {Y}_{LL}^{F}, {Y}_{LH}^{F}, {Y}_{HL}^{F}, {Y}_{HH}^{F})
\end{align}
where IAWT indicates the two-dimensional inverse adaptive wavelet transform and ${Y}^F$ refers to the filtered feature map.

\subsection{Implementation Details}
In our experiments, we employed the AdamW optimizer with a learning rate $10^{-4}$. 
The process used the PolyLR scheduler, the LinearLR warmup policy and the layer-wise learning rate decay policy. 
The batch size was 2.
This scheduling scheme was applied over 180,000 iterations. 
To effectively train the model proposed in this paper, we used a weighted loss consisting of the Dice loss and the binary cross-entropy (BCE) loss.
To initialize the network, the WTConv backbone was firstly pretrained on the ImageNet-1k classification dataset for over 300 epochs. 
The videos were temporally cropped starting from a randomized initial frame, resulting in many 10-frame sections along the front-to-back direction. 
The same parameters of data augmentations (e.g., random blur, random flip and color jitter) were then applied to all frames within each section.
A section was fed into the network in temporal order and the input images were resized to 512$\times$512. 
The model was implemented with PyTorch and mmsegmentation~\cite{mmseg2020}, and all the experiments were conducted on a single NVIDIA GeForce RTX 4090 GPU.
The reported speed is measured with a batch size of 1, and the performance is reported in milliseconds per-frame (ms/frame). 
The reported inference time corresponds to single-model, single-scale evaluation, and post-processing is excluded.
\section{Experiment and results}
\subsection{Datasets and Evaluation Metrics}
The methods were evaluated in four available ultrasound video datasets, including two thyroid nodules, a thyroid gland, a cardiac ultrasound datasets:
\textbf{The Thyroid Nodule dataset} consists of 64 ultrasound videos, totaling 5611 frames with pixelwise annotation of the thyroid nodules. The image resolution ranges from 333$\times$510 to 523 $\times$771. The dataset is split into 40, 10 and 14 for training, validation and testing, respectively.
\textbf{The VTUS dataset}~\cite{Vivim} comprises 100 ultrasound videos of thyroid ultrasound, totaling 9342 frames with pixel-wise ground truth.  The dataset includes transverse and longitudinal B-mode ultrasound videos captured by Mindray Resona8 and TOSHIBA Aplio500 vendors. These videos are cross-annotated by three experts with more than three years of experience in thyroid diagnosis. The number of frames in these videos ranges from 31 to 196. The entire dataset is partitioned into training and test sets at ratio of 7:3, resulting in 70 training and 30 testing videos.
\textbf{The TG3K dataset}~\cite{tg3k}
comprises 16 ultrasound thyroid videos with pixel-wise gland annotations.
Following Gong et al. \cite{ThyroidNodule1}, we selected the sequences in which the thyroid gland occupies more than 6\% of the entire image yielding 3,585 frames from 16 sequences with image resolutions varying from 277$\times$284 to 316$\times$333. Herein, 12 sequences were chosen for training, 2 were selected for validation, and 2 were selected for testing.
\textbf{The CAMUS datasets}~\cite{CAMUS} contains echocardiography data derived from 500 patients with pixelwise annotations of the left ventricle, these data were acquired at the University Hospital of St Etienne (France). The data consists of 1,000 standardized echocardiographic sequences: 500 apical 2-chamber views and 500 apical 4-chamber views. The 1000 sequences were allocated to training(700), validation(100)and test(200) sets.
\textbf{Different evaluation metrics} were employed in both comparative and ablation experiments, including the Dice similarity coefficient (DSC), intersection over union (IoU), mean absolute error (MAE), precision and recall metrics.

\subsection{Comparison with the SOTA Methods}

\begin{figure*}[!t]
	\centerline{\includegraphics[width=\textwidth]{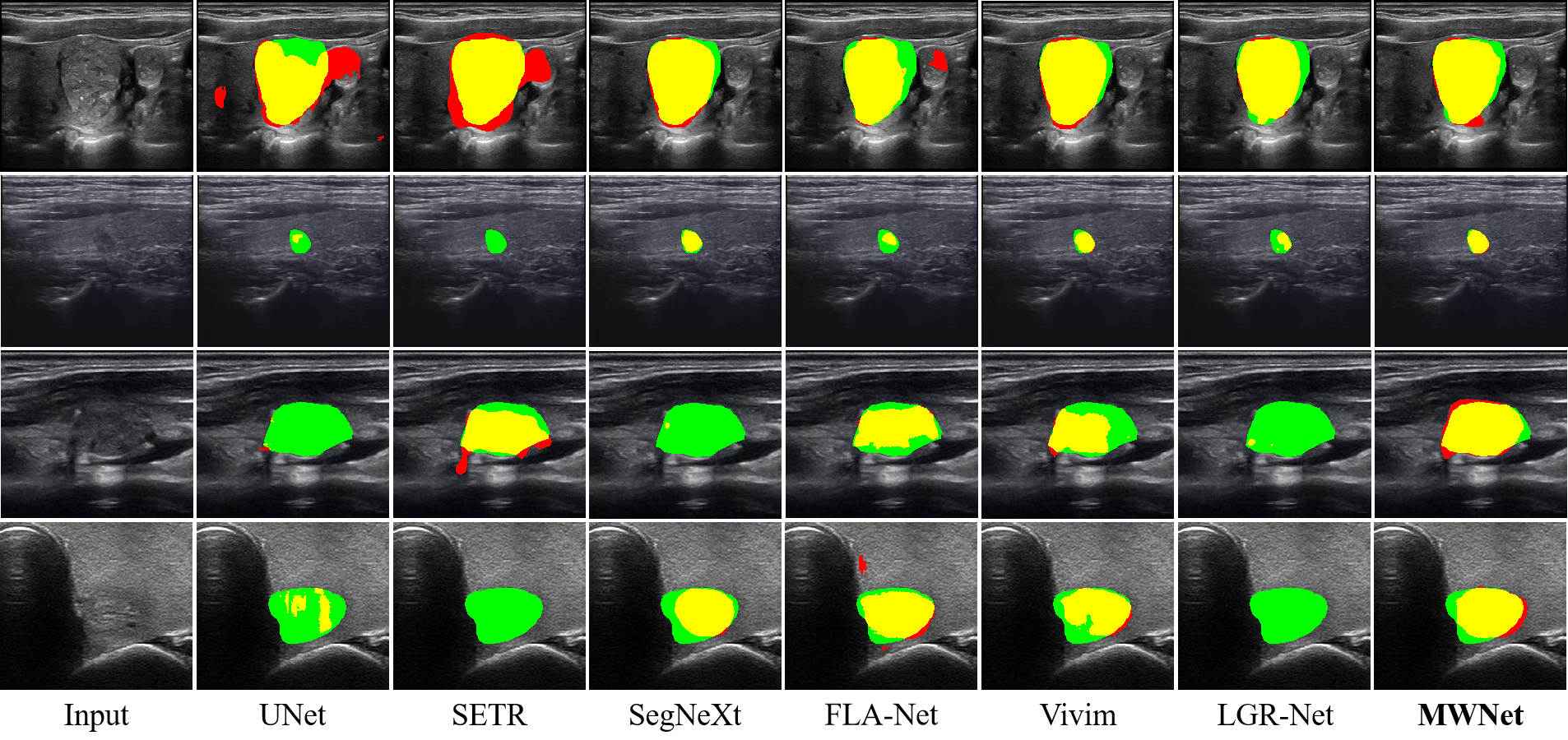}}
	\caption{The segmentation visualization results of the comparison experiments. Green, red, and yellow regions represent the ground truth, prediction, and their overlapping regions, respectively.  }
	\label{fig4}
\end{figure*}

Our proposed method was compared with the SOTA methods, including six approaches designed for images (UNet, Swin-UNet~\cite{SWinUnet}, SETR~\cite{SETR}, SegFormer~\cite{SegFormer}, SegNeXt~\cite{SegNeXt} and Mask2Former~\cite{Mask2Former}) and five methods for videos (FLA-Net~\cite{FLANet}, Vivim~\cite{Vivim}, LGR-Net~\cite{LGRNet}, OneVOS~\cite{OneVOS} and MemSAM~\cite{MemSAM}). The comparison results are shown in Table~\ref{tab1}, Table~\ref{tab2}, Table~\ref{tab3}, Table~\ref{tab4}. Our proposed MWNet achieved the highest DSC and IoU values, along with the lowest MAEs across all the datasets, demonstrating its robustness and superiority over the current SOTA methods.
In contrast, the other methods exhibited inferior segmentation accuracy due to their inherent limitations in terms of handling the aforementioned challenges. 
The ability of U-Net and FLA-Net to capture global contextual information is insufficient, especially when processing confusing locations.
For LGR-Net, the inadequate extraction of HF detail information resulted in reduced boundary segmentation accuracy, especially for small-scale objects. 
SETR and Vivim, which adopt feature pyramid networks (FPN)-like and MLP-based decoders~\cite{SegFormer}, respectively, suffered from insufficient feature fusion, leading to degraded segmentation performance, particularly at blurred boundaries. 
SegNext only utilized the last three layers of the feature maps, thus overlooking important low-level information. 
The above mentioned limitations were mitigated by our method to varying degrees. 
The visualization results on the Thyroid Nodule and VTUS datasets are shown in Fig.~\ref{fig4}. 
Specifically, the masks predicted by our method match the ground truth better and the boundary segmentation accuracy is notably improved, which validates the ability of our method to address the challenges of ultrasound video segmentation, including confusing locations, speckle noise, and blurred boundaries.
\begin{table}
	\caption{The results of methods in the test set of the Thyroid Nodule dataset}
	\label{table}
	\setlength{\tabcolsep}{2.5pt}	\begin{tabular}{p{50pt}p{30pt}p{30pt}p{25pt}p{25pt}p{30pt}p{15pt}}
		\hline
		Method & Type& DSC(\%)& IoU(\%)&MAE &\verb|#|param &FPS
		\\
		\hline
		UNet& Image& 75.88& 61.13&0.0184&28.99M&77\\
		Swin-UNet& Image& 75.73& 60.94&0.0208&27.17M&42\\
		SETR& Image& 79.60& 66.11& 0.0164&315.37M&27\\
		SegFormer& Image& 81.65& 68.99& 0.0139&44.6M&111\\
		SegNeXt& Image& 85.13& 74.11&0.0120&13.93M&91\\
		Mask2Former& Image& 83.91& 72.28& 0.0133&68.72M&48\\
		\hline
		FLA-Net& Video& 79.12& 65.46&0.0168&87.62M&37\\
		Vivim& Video& 83.57& 71.78&0.0116&57.7M&21\\
		LGR-Net& Video& 85.58& 74.79&0.0109&27.66M&62\\
		MemSAM& Video& 85.48& 74.64&0.0110&152.05M&45\\
		\hline
		MWNet& Video& $\mathbf{88.03}$& $\mathbf{78.61}$ &$\mathbf{0.0101}$& 72.99M&28\\
		\hline
	\end{tabular}
	\label{tab1}
\end{table}

\begin{table}
	\caption{The results of methods in the test set of the VTUS dataset}
	\label{table}
	\setlength{\tabcolsep}{3pt}	\begin{tabular}{p{50pt}p{30pt}p{30pt}p{30pt}p{37pt}p{30pt}}
		\hline
		Method & Type& DSC(\%)& IoU(\%) &MAE&\verb|#|param. \\
		\hline
		UNet& Image& 75.03& 60.04&0.0403&28.99M\\
		Swin-UNet& Image& 76.50& 61.94&0.0365&27.17M\\
		SETR& Image& 80.93& 67.97&0.0328&315.37M\\
		SegFormer& Image& 83.37& 71.48&0.0271&44.6M\\
		SegNeXt& Image& 83.05& 71.02&0.0277&13.93M\\
		Mask2Former& Image& 84.63& 73.35&0.0261&68.72M\\
		\hline
		FLA-Net& Video& 85.66& 74.92&0.0251&87.62M\\
		Vivim& Video& 86.22& 75.77&0.0240&57.7M\\
		LGR-Net& Video& 69.89& 53.71&0.0541&27.66M\\
		MemSAM& Video& 83.56& 71.76&0.0268&152.05M\\
		\hline
		MWNet& Video& $\mathbf{87.68}$& $\mathbf{78.06}$&$\mathbf{0.0208}$&72.99M\\
		\hline
	\end{tabular}
	\label{tab2}
\end{table}
We employ the t-test for the IoU to demonstrate the performance differences between our method and other SOTA methods.
The $p$ value of MWNet compared with SOTA methods is $p<0.05$ in the Thyroid Nodule dataset and the VTUS dataset. 
These results indicate that our method achieves superior performance compared with existing approaches and offers distinct advantages in small object segmentation. 
The $p$ value in the TG3K dataset is $p<0.05$ except for SegFormer ($p = 0.43$) and SegNeXt ($p = 0.07$). 
The results show minimal performance differences between video-based and single-frame processing. 
It may be attributed to the limited variation between frames in thyroid ultrasound videos. 
The existing ultrasound video segmentation methods primarily focus on adjacent or short-term frame features, without fully leveraging long-range temporal dependencies. 
In the CAMUS dataset, except for Vivim ($p = 0.41$) and LGR-Net ($p = 0.53$), the $p$ value is $p<0.05$. 
This phenomenon may result from the large and well-defined objects in echocardiography, where segmentation is inherently easier and performance differences among methods become less pronounced.
\begin{table}
	\caption{The results of methods in the test set of the TG3K dataset}
	\label{table}
	\setlength{\tabcolsep}{3pt}	\begin{tabular}{p{50pt}p{37pt}p{30pt}p{30pt}p{30pt}p{30pt}}
		\hline
		Method & Type& DSC(\%)& IoU(\%)& MAE&\verb|#|param.\\
		\hline
		UNet& Image& 86.27& 75.85&  0.0288&28.99M\\
		Swin-UNet& Image& 85.73& 75.03& 0.0296&27.17M\\
		SegFormer& Image& 87.51& 77.80& 0.0265&44.6M\\
		SegNeXt& Image& 87.55& 77.85& 0.0270&13.93M\\
		\hline
		FLA-Net& Video& 83.32& 71.41& 0.0322&87.62M\\
		Vivim& Video& 87.37& 77.57&0.0259&57.7M\\
		LGR-Net& Video& 87.12& 77.18&0.0273&27.66M\\
		MemSAM& Video& 86.39& 76.04& 0.0280&152.05\\
		\hline
		MWNet& Video& $\mathbf{87.90}$& $\mathbf{78.41}$& $\mathbf{0.0258}$&72.99M\\
		\hline
		\multicolumn{6}{p{203pt}}{}
	\end{tabular}
	\label{tab3}
\end{table}
\begin{table}
	\caption{The results of methods in the test set of the CAMUS dataset}
	\label{table}
	\setlength{\tabcolsep}{3pt}	\begin{tabular}{p{50pt}p{37pt}p{30pt}p{30pt}p{30pt}p{30pt}}
		\hline
		Method & Type& DSC(\%)& IoU(\%)& MAE&\verb|#|param.\\
		\hline
		UNet& Image& 93.37& 87.57&  0.0127&28.99M\\
		Swin-UNet& Image& 93.09& 87.07& 0.0132&27.17M\\
		SegFormer& Image& 93.14& 87.15& 0.0135&44.6M\\
		SegNeXt& Image& 93.12& 87.12& 0.0131&13.93M\\
		\hline
		FLA-Net& Video& 94.01& 88.71& 0.0114&87.62M\\
		Vivim& Video& 94.19& 89.02&0.0112&57.7M\\
		LGR-Net& Video& 94.06& 88.79&0.0114&27.66M\\
		MemSAM& Video& 93.77& 88.27& 0.0119&152.05M\\
		\hline
		MWNet& Video& $\mathbf{94.34}$& $\mathbf{89.29}$& $\mathbf{0.0109}$&72.99M\\
		\hline
		\multicolumn{6}{p{203pt}}{}
	\end{tabular}
	\label{tab4}
\end{table}
\subsection{Ablation Experiments}
In the ablation experiments, WTConv backbone was replaced with a standard ConvNeXt~\cite{ConvNeXt}, the temporal fusion module and the memory bank were removed, and the HFF module was replaced by the upsampling method in UNet. 
The comparison results are shown in Table~\ref{tab5}.
The addition of the above modules (WTConv in encoder, memory module and HFF in decoder) resulted in improvements in the DSC metric by 0-1.83\%, 1.8-3.34\%, and 0.24-0.82\%, respectively. 
The inclusion of memory modules yields notable improvements in segmentation performance, emphasizing the importance of inter-frame temporal information. 
The larger gains observed on the thyroid nodule dataset further indicate that small object segmentation is particularly affected by ultrasound video limitations, leading to higher risks of tracking loss.  
Specifically, the MWConv encoder could effectively model spatiotemporal features by integrating local details with global dependencies, thus obtaining multiscale and multiresolution feature maps.
And the memory mechanism enables object tracking by modeling both long-term and short-term temporal dependencies.
Additionally, the HFF module significantly improves the accuracy of boundary segmentation, especially for small objects, by fusing HF detail information acquired from shallow feature maps with LF semantic information derived from deep feature maps.
The importance and detailed behavior of the proposed backbone and fusion module are further examined in the Discussion section.

\begin{table}
	\caption{The ablation experiment of MWNet on four datasets}
	\label{table}
	\setlength{\tabcolsep}{2pt}
	\begin{tabular}{p{10pt}p{33pt}p{35pt}p{20pt}p{20pt}p{20pt}p{34pt}p{34pt}}
		\hline
		&WTConv& Memory& HFF& DSC& IoU& Precision&Recall\\
		\hline
		\multirow{4}{*}{A}&\ding{56}& \ding{56}& \ding{56}& 82.63& 70.40& 89.76&76.55\\
		&\ding{52}& \ding{56}& \ding{56}& 84.46&73.10&$\mathbf{90.8}$&78.95\\
		&\ding{52}& \ding{52}& \ding{56}& 87.21&77.32&90.73&83.95\\
		&\ding{52}& \ding{52}& \ding{52}& $\mathbf{88.03}$&$\mathbf{78.61}$&90.04&$\mathbf{86.1}$\\
		\hline
		\multirow{4}{*}{B}&\ding{56}& \ding{56}& \ding{56}& 82.24& 69.84& 86.24&78.6\\
		&\ding{52}& \ding{56}& \ding{56}& 83.67&71.92&89.56&78.5\\
		&\ding{52}& \ding{52}& \ding{56}& 87.01&77.01&$\mathbf{91.57}$&82.89\\
		&\ding{52}& \ding{52}& \ding{52}& $\mathbf{87.68}$&$\mathbf{78.06}$&88.83&$\mathbf{86.56}$\\
		\hline
		\multirow{4}{*}{C}&\ding{56}& \ding{56}& \ding{56}& 86.9& 76.84& 90.05&83.97\\
		&\ding{52}& \ding{56}& \ding{56}& 87.55&77.85&89.9&85.32\\
		&\ding{52}& \ding{52}& \ding{56}& 87.65&78.01&$\mathbf{90.06}$&85.37\\
		&\ding{52}& \ding{52}& \ding{52}& $\mathbf{87.9}$&$\mathbf{78.41}$&89.24&$\mathbf{86.57}$\\
		\hline
		\multirow{4}{*}{D}&\ding{56}& \ding{56}& \ding{56}& 92.38& 86.12& 91.97&92.85\\
		&\ding{52}& \ding{56}& \ding{56}& 92.3&85.99&92.03&92.6\\
		&\ding{52}& \ding{52}& \ding{56}& 94.1&88.86&$\mathbf{93.93}$&94.26\\
		&\ding{52}& \ding{52}& \ding{52}& $\mathbf{94.34}$&$\mathbf{89.29}$&93.64&$\mathbf{95.05}$\\
		\hline
		\multicolumn{8}{p{230pt}}{A, B, C, D represent the Thyroid Nodule, VTUS, TG3K, CAMUS datasets. \ding{52} represents this module used in MWNet and \ding{56} indicates abandoned module. 'Memory' includes the temporal fusion module and the memory bank.}\end{tabular}
	\label{tab5}
\end{table}

\section{Discussion}
\label{discussion}
\begin{figure}[!t]
	\centerline{\includegraphics[scale=0.25]{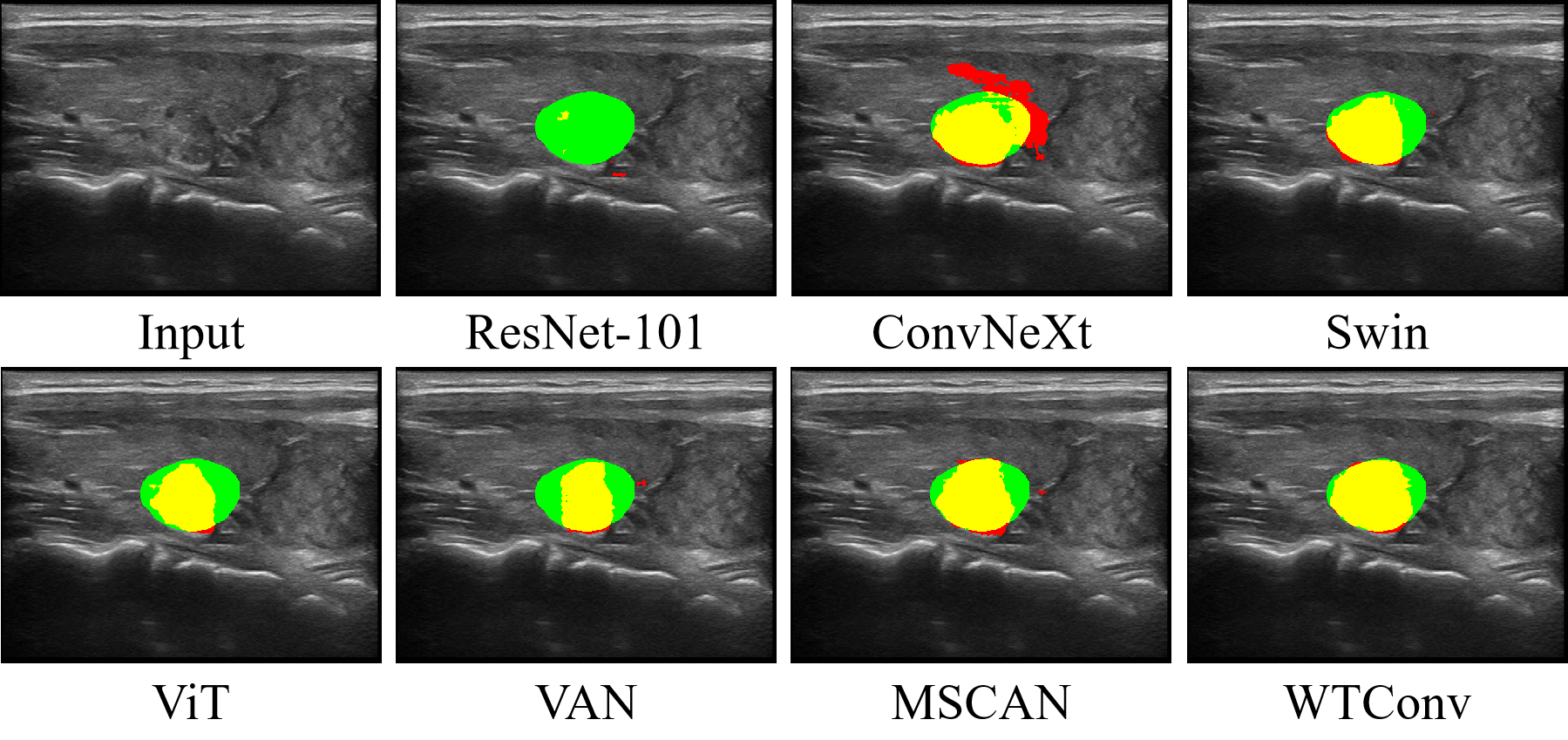}}
	\caption{The segmentation visualization results of comparison backbones on the Thyroid Nodule Dataset.  Red, green, and yellow regions represent the ground truth, prediction, and their overlapping regions, respectively.}
	\label{fig5}
\end{figure}

\begin{table}
	\caption{Comparison with state-of-art backbones on Thyroid Nodule dataset}
	\label{table}
	\setlength{\tabcolsep}{2pt}
	\begin{tabular}{p{100pt}p{30pt}p{30pt}p{33pt}p{33pt}}
		\hline
		Backbone& DSC(\%)& IoU(\%)& Precision&Recall\\
		\hline
		ResNet& 79.52& 66.01& 89.02&71.85\\
		ConvNeXt\cite{ConvNeXt}& 84.75& 73.53& $\mathbf{89.75}$&80.27\\
		ViT& 82.73&70.55&88.96&77.32\\
		Swin-T\cite{SwinTransformer}& 82.08& 69.60& 89.43&75.84\\
		VAN\cite{VAN}& 82.94& 70.85& 89.43&77.32\\
		MSCAN\cite{SegNeXt}& 85.43& 74.56& 88.94&82.18\\
		\hline
		WTConv& $\mathbf{85.75}$& $\mathbf{75.06}$& 89.65&$\mathbf{82.18}$\\
		\hline
	\end{tabular}
	\label{tab6}
\end{table}

\begin{figure}[!t]
	\centerline{\includegraphics[scale=0.25]{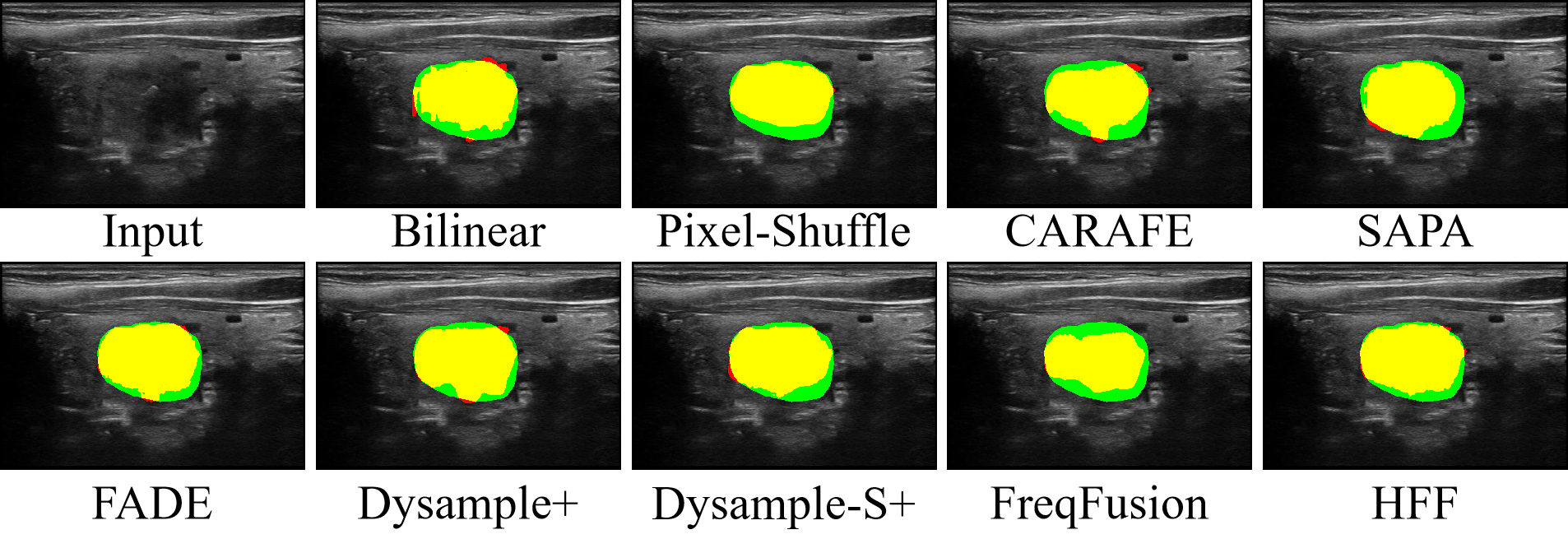}}
	\caption{The segmentation visualization results of comparison feature fusion methods on the Thyroid Nodule Dataset.  Red, green, and yellow regions represent the ground truth, prediction, and their overlapping regions, respectively.}
	\label{fig6}
\end{figure}

\begin{table}
	\caption{Comparison with state-of-the-art feature fusion methods on Thyroid Nodule dataset}
	\label{table}
	\setlength{\tabcolsep}{2pt}
	\begin{tabular}{p{80pt}p{30pt}p{30pt}p{33pt}p{30pt}}
		\hline
		Fusion Method& DSC(\%)& IoU(\%)& Precision&Recall\\
		\hline
		Nearest& 84.46& 73.10& 90.80&78.95\\
		Bilinear& 84.80&73.61&90.25&79.97\\
		Pixel-Shuffle~\cite{PixelShuffle}& 84.11&72.58&90.60&78.49\\
		CARAFE~\cite{CARAFE}& 85.32&74.40&$\mathbf{92.32}$&79.31\\
		SAPA~\cite{SAPA}& 84.73& 73.50& 91.20&79.11\\
		FADE~\cite{FADE}& 85.24& 74.28& 91.43&79.84\\
		Dysample+~\cite{Dysample}& 85.15& 74.14& 89.37&81.32\\
		Dysample-S+~\cite{Dysample}& 84.70& 73.46& 89.96&80.02\\
		FreqFusion~\cite{FreqFusion}& 85.18& 74.19& 90.59&80.38\\
		\hline
		HFF& $\mathbf{85.75}$& $\mathbf{75.06}$& 89.65&$\mathbf{82.18}$\\
		\hline
	\end{tabular}
	\label{tab7}
\end{table}

\begin{figure}[!t]
	\centerline{\includegraphics[scale=0.3]{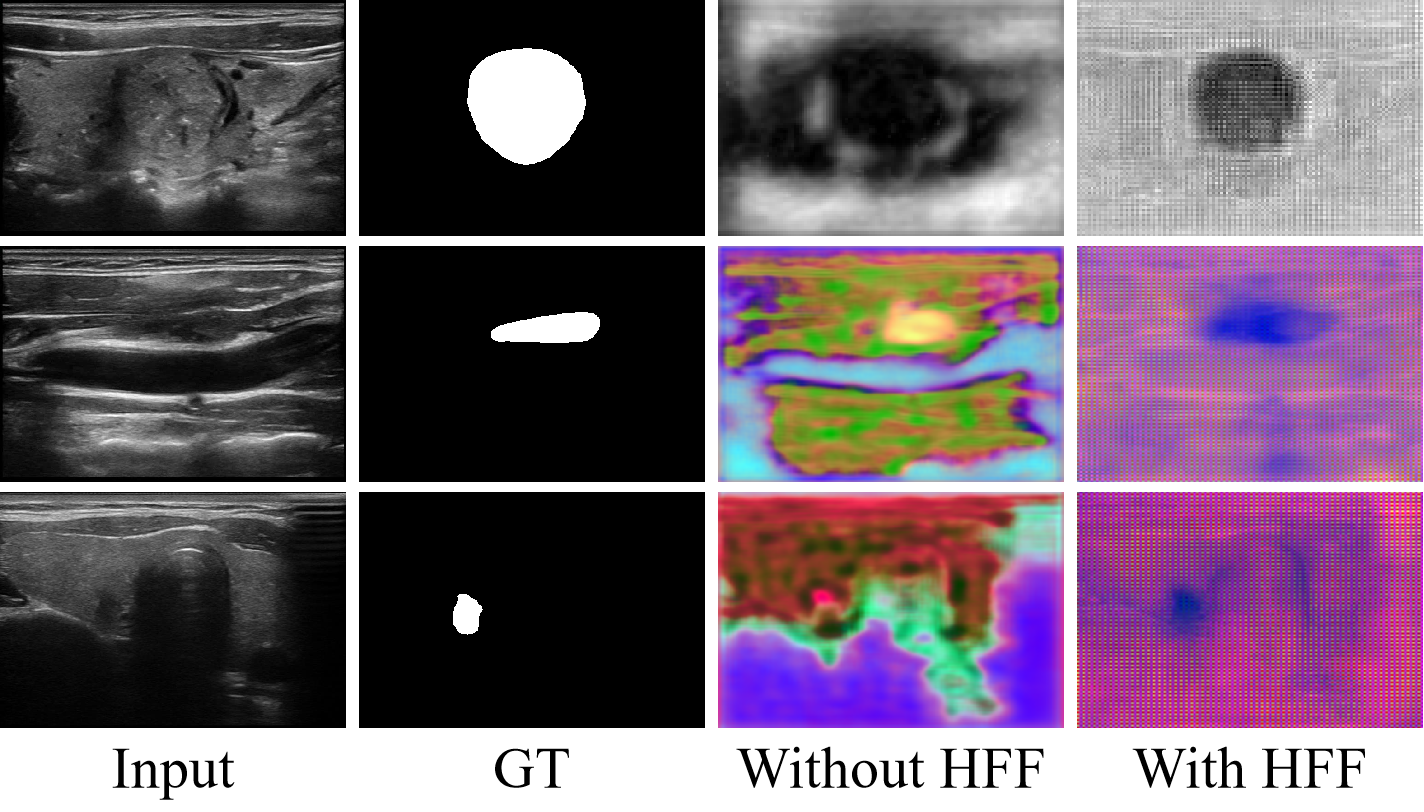}}
	\caption{The visualization results of the shallowest feature maps in the decoder. The first row presents the visualization of a single feature map. The next two rows show BGR images generated by mapping the first three principal components after applying principal component analysis to the feature map. "Without HFF" denotes replacing the decoder module with a standard interpolation-based upsampling module.}
	\label{fig7}
\end{figure}

\subsection{Comparison between WTConv backbone and SOTA backbones}
The classical backbones with small version were selected for comparison with our proposed WTConv backbone, which was produced by removing the temporal fusion module from the MWConv encoder. 
The videos in the Thyroid Nodule dataset were split into individual frames that served as network inputs, and a comparison among the produced evaluation metrics is shown in Table~\ref{tab6}. 
ConvNeXt and MSCAN, benefited from the small neighborhood convolution and the large receptive field of its kernels, achieving performance that was second only to that of our proposed method. 
Notably, the utilization of multiscale features by MSCAN yielded a marked improvement, motivating us to incorporate frequency-domain decomposition into our approach.
In contrast, the ViT, Swin-T and VAN emphasize capturing long-distance dependencies, which limits their effectiveness at handling low-contrast ultrasound videos. 
Inspired by them, we integrate wavelet analysis theory into the convolution operation to increase its sensitivity to HF variations, improving detailed feature extraction without sacrificing the receptive field and facilitating the generation of multiresolution, multiscale feature representations. 
The visualization results of the comparison are shown in Fig.~\ref{fig5}. 
Our predicted masks demonstrate the highest consistency with the ground-truth annotations and fewer false segmentation.
This indicates the effectiveness of our method for accurately recognizing and segmenting small object boundaries.
In summary, the experimental results indicate that WTConv backbone has the potential to be a generalized backbone for ultrasound image/video processing tasks.

\subsection{Comparison of HFF with other feature fusion module}
The HFF module was replaced with nine fusion methods, which are shown in Table~\ref{tab7}. 
Our method achieved the highest DSC, IoU, and recall scores, demonstrating the most accurate alignment with the ground-truth masks, especially in the foreground regions. 
Among these SOTA methods, CARAFE, FADE, Dysample+, and FreqFusion also achieve strong performance, ranking just below our proposed method. 
CARAFE and FADE benefited from capturing high-resolution structures. 
Dysample+ mitigated inconsistencies in potential features using a refined upsampling strategy. 
FreqFusion combined the advantages of both low-level and high-level feature representations via frequency-aware fusion and adaptive resampling to effectively integrate global and local features. Inspired by these methods and aiming to increase the utilization rate of HF information, we proposed the HFF module, which integrates adaptive wavelets that are both HF-sensitive and noise-unaware to facilitate the efficient fusion of multiresolution features, with a particular emphasis on high-resolution detail components, ultimately leading to optimal performance. 
The typical visualization results are shown in Fig.~\ref{fig6}.
To demonstrate the effectiveness of the HFF module in terms of addressing boundary displacement issues in ultrasound segmentation tasks, the fused feature maps generated from both bilinear interpolation and the proposed HFF module were visualized, as shown in Fig.~\ref{fig7}. 
The first row presents the visualization of a single feature map. 
The next two rows show BGR images generated by mapping the first three principal components after applying principal component analysis to the feature map. 
It can be seen that our method produces clearer boundary delineation and cleaner background regions.
The HFF module could effectively suppress the interfering HF components using the AWLF mechanism, thereby reducing noise and intraclass inconsistency. 
Furthermore, it could enhance the relevant HF features through the AWHF mechanism, resulting in sharper and more accurate boundary representations.

\subsection{Effectiveness of adaptive wavelet transform}

\begin{table}
	\caption{Comparison with wavlet basis on Thyroid Nodule dataset}
	\label{table}
	\setlength{\tabcolsep}{2pt}
	\begin{tabular}{p{70pt}p{32pt}p{32pt}p{34pt}p{32pt}}
		\hline
		Wavelet Basis& DSC(\%)& IoU(\%)& Precision&Recall\\
		\hline
		Haar& 85.05& 66.01& 89.02&71.85\\
		Daubechies
		& 85.52&70.55&88.96&77.32\\
		Symlet& 82.08& 69.60& 89.43&75.84\\
		\hline
		Adaptive& $\mathbf{85.75}$& $\mathbf{75.06}$& $\mathbf{89.65}$&$\mathbf{82.18}$\\
		\hline
	\end{tabular}
	\label{tab8}
\end{table}
The wavelet basis is a major factor for the effectiveness of wavelet transform.
However, a fixed wavelet basis may not be universally applicable across different layers of feature maps, due to their distinct characteristics and the diverse requirements of various tasks. 
By substituting the AWT module with conventional wavelet transforms featuring fixed wavelet bases, including the Haar, Daubechies, Symlet, and Coiflet wavelets, we verified the effectiveness of adaptive wavelet fusion. 
As demonstrated in Table~\ref{tab8}, the HFF module employing the AWT yielded the optimal results, which demonstrates the superior performance and flexibility of the AWT.

\subsection{Effectiveness of long short-term memory bank}
\begin{figure}[!t]
	\centerline{\includegraphics[scale=0.4]{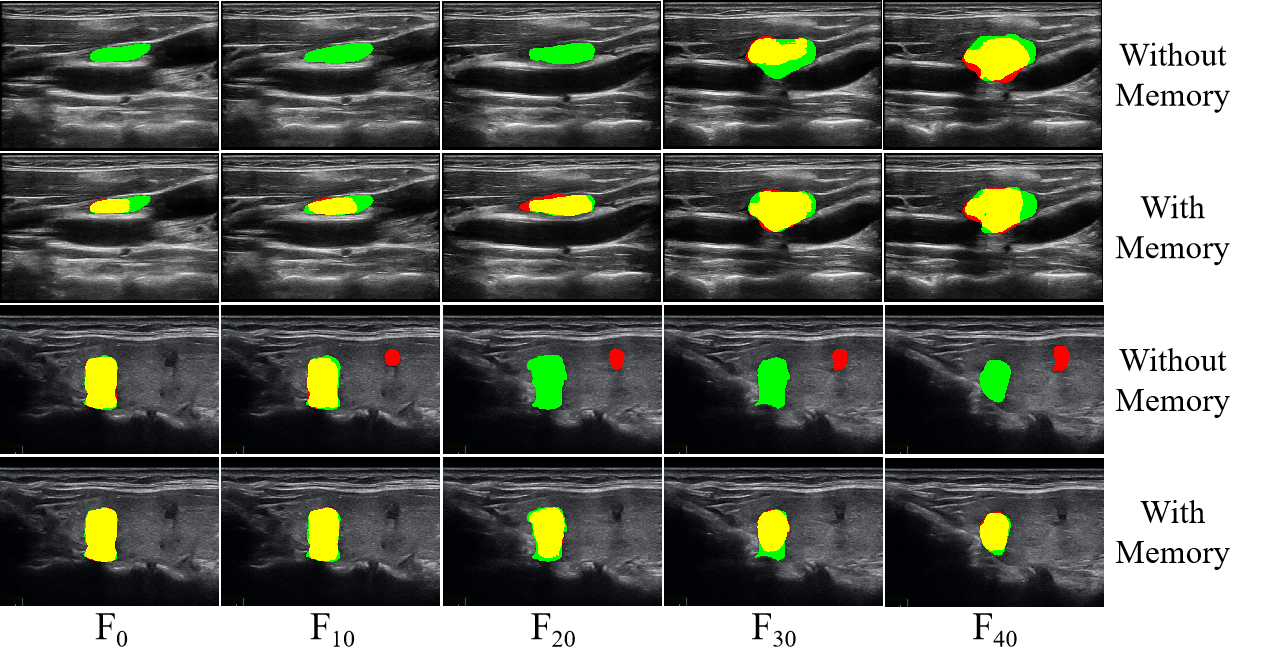}}
	\caption{The visualization results of the visualization results with ten frames interval  for adding (with memory) and removing (without memory) long short-term memory bank.  Green, red, and yellow regions represent the ground truth, prediction, and their overlapping regions, respectively. F$_0$, F$_{10}$, F$_{20}$, F$_{30}$, F$_{40}$ represent the prediction results for frame 0, 10, 20, 30, 40.}
	\label{fig9}
\end{figure}

\begin{figure}[!t]
	\centerline{\includegraphics[scale=0.4]{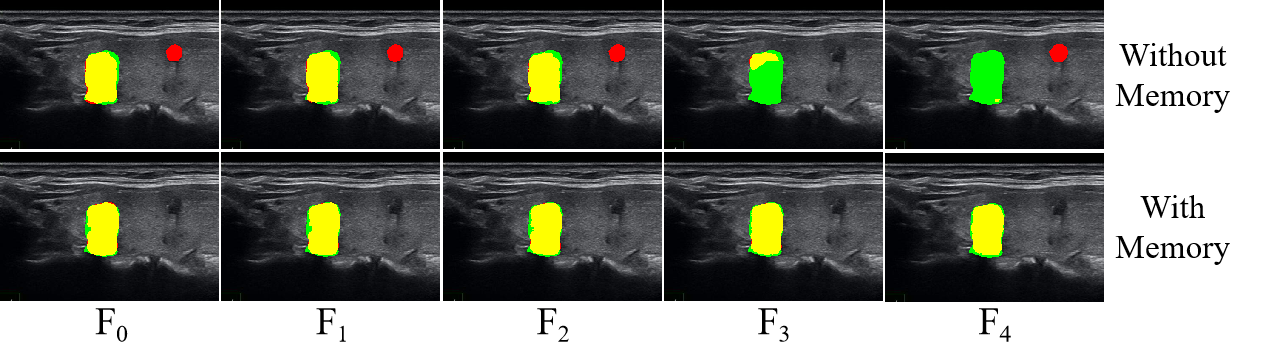}}
	\caption{The visualization results of adjacent five frames for adding (with memory) and removing (without memory) long short-term memory bank. Green, red, and yellow regions represent the ground-truth, prediction, and their overlapping regions, respectively. F$_0$, F$_1$, F$_2$, F$_3$, F$_4$ represent the prediction results for frame 0, 1, 2, 3, 4.}
	\label{fig10}
\end{figure}

The visualization results with ten frames interval for adding (with memory) and removing (without memory) long short-term memory bank are shown in Fig.~\ref{fig9}, highlighting the enhanced temporal tracking capability for long video.
It indicates that adding the memory bank can effectively improve blurred boundaries and confusing locations, especially long-range segmentation object tracking.
The visualization results of adjacent five frames for adding (with memory) and removing (without memory) long short-term memory bank are shown in Fig.~\ref{fig10}. 
It can be observed that the results without memory exhibit noticeable boundary jitter, whereas the inclusion of memory effectively maintains temporal consistency and mitigates large segmentation discrepancies between adjacent frames. 

\begin{figure}[!t]
	\centerline{\includegraphics[scale=0.2]{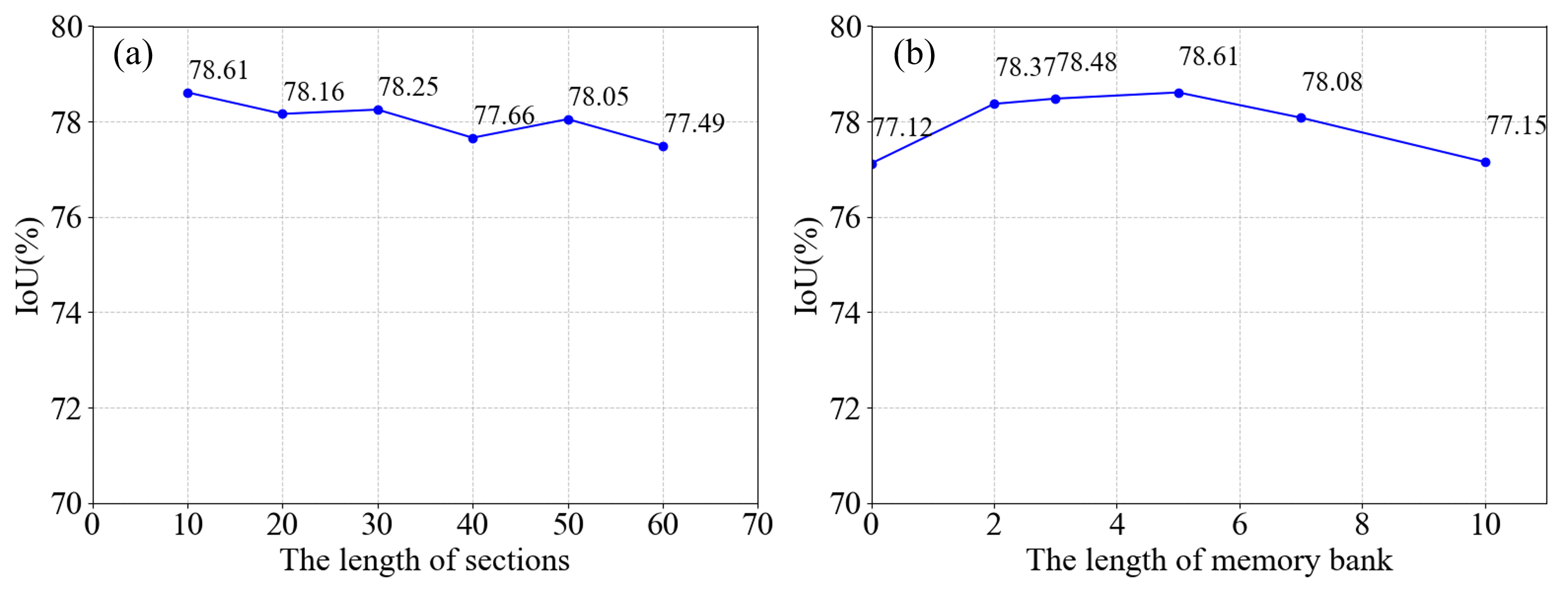}}
	\caption{The relationship between section length (or memory bank length) and segmentation precision. (a) section length. The memory bank length is 5. (b) memory bank length. The section length is 10.}
	\label{fig11}
\end{figure}

We drew inspiration from previous ultrasound video segmentation studies~\cite{MemSAM,Vivim}, which commonly used section lengths of 5 and 10. 
Accordingly, the length was adopted for 10 in this work and then the accuracy across different lengths was compared. 
The lengths of the long-term and short-term memory banks are set to 5 and 2, respectively. 
And the long videos are divided into multiple frame sections with fixed lengths (10, 20, 30, 40, 50, 60). 
It is shown in Fig.~\ref{fig11}(a). 
The results indicate that the segmentation accuracy remains largely stable across different section lengths, confirming the adaptability and robustness of our proposed memory mechanism.
We added a comparison of different memory bank lengths, as shown in Fig.~\ref{fig11}(b). 
It can be observed that as the length increases, segmentation precision initially improves and then declines. 
It indicates that a restricted memory bank can effectively mitigate redundancy introduced by mechanically storing all historical information. 

Overall, we have validated the effectiveness of our method on four ultrasound datasets: two thyroid nodule datasets, one thyroid dataset, and one echocardiography dataset. 
The cross-dataset comparison demonstrates the adaptability of our method to diverse ultrasound video types, with particularly notable improvements on thyroid nodule datasets. 
This can be attributed to the small nodule sizes, which exacerbate the inherent challenges of ultrasound image segmentation and increase the likelihood of object loss for video tracking. 
In contrast, performance improvements comparison with SOTA methods are less evident for larger objects, such as the thyroid and cardiac structures, which show minimal inter-frame variation and are inherently easier to segment. 
Our method can effectively address challenges such as blurred boundaries, confusing locations, speckle noise in ultrasound videos and shows superior performance for small object in long-duration ultrasound sequences.

The prediction speed results in Table~\ref{tab1} indicate that video-based segmentation methods typically incur higher computational overhead than single-frame processing, which can be attributed to the incorporation of temporal feature modeling.
Although the proposed method demonstrates superior performance over existing SOTA approaches, its prediction efficiency and GPU memory utilization remain suboptimal, primarily due to the inclusion of the memory module. 
In future work, we plan to further investigate optimized memory-based module to enhance segmentation robustness and efficiency in long ultrasound videos. 
Moreover, we will explore the generalization of the proposed framework to other medical long video.
\section{Conclusion}
In this work, we propose MWNet to effectively extract multi-scale feature maps in wavelet domain and emphasize the boundary-sensitive HF details, then achieving highly accurate ultrasound video segmentation.
The proposed MWConv, memory bank, HFF module can be employed as a generalized backbone and feature fusion module in segmentation tasks, especially for low-quality images and small object in long video.
Extensive experiments conducted on four datasets validate the superiority and generalization of our proposed model.

\printcredits
\bibliographystyle{cas-model2-names}

\bibliography{reference}

@inproceedings{SWinUnet,
  title={Swin-unet: Unet-like pure transformer for medical image segmentation},
  author={Cao, Hu and Wang, Yueyue and Chen, Joy and Jiang, Dongsheng and Zhang, Xiaopeng and Tian, Qi and Wang, Manning},
  booktitle={European conference on computer vision},
  pages={205--218},
  year={2022},
  organization={Springer}
}

@inproceedings{SETR,
  title={Rethinking semantic segmentation from a sequence-to-sequence perspective with transformers},
  author={Zheng, Sixiao and Lu, Jiachen and Zhao, Hengshuang and Zhu, Xiatian and Luo, Zekun and Wang, Yabiao and Fu, Yanwei and Feng, Jianfeng and Xiang, Tao and Torr, Philip HS and others},
  booktitle={Proceedings of the IEEE/CVF conference on computer vision and pattern recognition},
  pages={6881--6890},
  year={2021}
}

@article{SegFormer,
  title={SegFormer: Simple and efficient design for semantic segmentation with transformers},
  author={Xie, Enze and Wang, Wenhai and Yu, Zhiding and Anandkumar, Anima and Alvarez, Jose M and Luo, Ping},
  journal={Advances in neural information processing systems},
  volume={34},
  pages={12077--12090},
  year={2021}
}

@article{SegNeXt,
  title={Segnext: Rethinking convolutional attention design for semantic segmentation},
  author={Guo, Meng-Hao and Lu, Cheng-Ze and Hou, Qibin and Liu, Zhengning and Cheng, Ming-Ming and Hu, Shi-Min},
  journal={Advances in neural information processing systems},
  volume={35},
  pages={1140--1156},
  year={2022}
}

@inproceedings{Mask2Former,
  title={Masked-attention mask transformer for universal image segmentation},
  author={Cheng, Bowen and Misra, Ishan and Schwing, Alexander G and Kirillov, Alexander and Girdhar, Rohit},
  booktitle={Proceedings of the IEEE/CVF conference on computer vision and pattern recognition},
  pages={1290--1299},
  year={2022}
}

@inproceedings{FLANet,
  title={Shifting more attention to breast lesion segmentation in ultrasound videos},
  author={Lin, Junhao and Dai, Qian and Zhu, Lei and Fu, Huazhu and Wang, Qiong and Li, Weibin and Rao, Wenhao and Huang, Xiaoyang and Wang, Liansheng},
  booktitle={International Conference on Medical Image Computing and Computer-Assisted Intervention},
  pages={497--507},
  year={2023},
  organization={Springer}
}

@article{Vivim,
  title={Vivim: a Video Vision Mamba for Ultrasound Video Segmentation},
  author={Yang, Yijun and Xing, Zhaohu and Yu, Lequan and Fu, Huazhu and Huang, Chunwang and Zhu, Lei},
  journal={IEEE Transactions on Circuits and Systems for Video Technology},
  year={2025},
  publisher={IEEE}
}

@inproceedings{LGRNet,
  title={Lgrnet: Local-global reciprocal network for uterine fibroid segmentation in ultrasound videos},
  author={Xu, Huihui and Yang, Yijun and Aviles-Rivero, Angelica I and Yang, Guang and Qin, Jing and Zhu, Lei},
  booktitle={International Conference on Medical Image Computing and Computer-Assisted Intervention},
  pages={667--677},
  year={2024},
  organization={Springer}
}

@inproceedings{SwinTransformer,
  title={Swin transformer: Hierarchical vision transformer using shifted windows},
  author={Liu, Ze and Lin, Yutong and Cao, Yue and Hu, Han and Wei, Yixuan and Zhang, Zheng and Lin, Stephen and Guo, Baining},
  booktitle={Proceedings of the IEEE/CVF international conference on computer vision},
  pages={10012--10022},
  year={2021}
}

@inproceedings{ConvNeXt,
  title={A convnet for the 2020s},
  author={Liu, Zhuang and Mao, Hanzi and Wu, Chao-Yuan and Feichtenhofer, Christoph and Darrell, Trevor and Xie, Saining},
  booktitle={Proceedings of the IEEE/CVF conference on computer vision and pattern recognition},
  pages={11976--11986},
  year={2022}
}

@article{VAN,
  title={Visual attention network},
  author={Guo, Meng-Hao and Lu, Cheng-Ze and Liu, Zheng-Ning and Cheng, Ming-Ming and Hu, Shi-Min},
  journal={Computational visual media},
  volume={9},
  number={4},
  pages={733--752},
  year={2023},
  publisher={Springer}
}

@inproceedings{WTConv,
  title={Wavelet convolutions for large receptive fields},
  author={Finder, Shahaf E and Amoyal, Roy and Treister, Eran and Freifeld, Oren},
  booktitle={European Conference on Computer Vision},
  pages={363--380},
  year={2024},
  organization={Springer}
}

@inproceedings{DAWN,
  title={Deep adaptive wavelet network},
  author={Rodriguez, Maria Ximena Bastidas and Gruson, Adrien and Polania, Luisa and Fujieda, Shin and Prieto, Flavio and Takayama, Kohei and Hachisuka, Toshiya},
  booktitle={Proceedings of the IEEE/CVF Winter Conference on Applications of Computer Vision},
  pages={3111--3119},
  year={2020}
}

@article{LiftingScheme,
  title={The lifting scheme: A construction of second generation wavelets},
  author={Sweldens, Wim},
  journal={SIAM journal on mathematical analysis},
  volume={29},
  number={2},
  pages={511--546},
  year={1998},
  publisher={SIAM}
}

@inproceedings{PixelShuffle,
  title={Real-time single image and video super-resolution using an efficient sub-pixel convolutional neural network},
  author={Shi, Wenzhe and Caballero, Jose and Husz{\'a}r, Ferenc and Totz, Johannes and Aitken, Andrew P and Bishop, Rob and Rueckert, Daniel and Wang, Zehan},
  booktitle={Proceedings of the IEEE conference on computer vision and pattern recognition},
  pages={1874--1883},
  year={2016}
}

@inproceedings{CARAFE,
  title={Carafe: Content-aware reassembly of features},
  author={Wang, Jiaqi and Chen, Kai and Xu, Rui and Liu, Ziwei and Loy, Chen Change and Lin, Dahua},
  booktitle={Proceedings of the IEEE/CVF international conference on computer vision},
  pages={3007--3016},
  year={2019}
}

@inproceedings{FADE,
  title={FADE: Fusing the assets of decoder and encoder for task-agnostic upsampling},
  author={Lu, Hao and Liu, Wenze and Fu, Hongtao and Cao, Zhiguo},
  booktitle={European Conference on Computer Vision},
  pages={231--247},
  year={2022},
  organization={Springer}
}

@article{SAPA,
  title={SAPA: Similarity-aware point affiliation for feature upsampling},
  author={Lu, Hao and Liu, Wenze and Ye, Zixuan and Fu, Hongtao and Liu, Yuliang and Cao, Zhiguo},
  journal={Advances in Neural Information Processing Systems},
  volume={35},
  pages={20889--20901},
  year={2022}
}

@inproceedings{Dysample,
  title={Learning to upsample by learning to sample},
  author={Liu, Wenze and Lu, Hao and Fu, Hongtao and Cao, Zhiguo},
  booktitle={Proceedings of the IEEE/CVF International Conference on Computer Vision},
  pages={6027--6037},
  year={2023}
}

@article{FreqFusion,
  title={Frequency-aware feature fusion for dense image prediction},
  author={Chen, Linwei and Fu, Ying and Gu, Lin and Yan, Chenggang and Harada, Tatsuya and Huang, Gao},
  journal={IEEE Transactions on Pattern Analysis and Machine Intelligence},
  year={2024},
  publisher={IEEE}
}

@inproceedings{DPSTT,
  title={Rethinking breast lesion segmentation in ultrasound: A new video dataset and a baseline network},
  author={Li, Jialu and Zheng, Qingqing and Li, Mingshuang and Liu, Ping and Wang, Qiong and Sun, Litao and Zhu, Lei},
  booktitle={International Conference on Medical Image Computing and Computer-Assisted Intervention},
  pages={391--400},
  year={2022},
  organization={Springer}
}

@inproceedings{MemSAM,
  title={MemSAM: taming segment anything model for echocardiography video segmentation},
  author={Deng, Xiaolong and Wu, Huisi and Zeng, Runhao and Qin, Jing},
  booktitle={Proceedings of the IEEE/CVF Conference on Computer Vision and Pattern Recognition},
  pages={9622--9631},
  year={2024}
}

@article{ConvLSTM,
  title={Convolutional LSTM network: A machine learning approach for precipitation nowcasting},
  author={Shi, Xingjian and Chen, Zhourong and Wang, Hao and Yeung, Dit-Yan and Wong, Wai-Kin and Woo, Wang-chun},
  journal={Advances in neural information processing systems},
  volume={28},
  year={2015}
}

@article{RMFG_net,
  title={Refined feature-based Multi-frame and Multi-scale Fusing Gate network for accurate segmentation of plaques in ultrasound videos},
  author={Hu, Xifeng and Cao, Yankun and Hu, Weifeng and Zhang, Wenzhen and Li, Jing and Wang, Chuanyu and Mukhopadhyay, Subhas Chandra and Li, Yujun and Liu, Zhi and Li, Shuo},
  journal={Computers in Biology and Medicine},
  volume={163},
  pages={107091},
  year={2023},
  publisher={Elsevier}
}

@article{ThyroidNodule1,
  title={Thyroid region prior guided attention for ultrasound segmentation of thyroid nodules},
  author={Gong, Haifan and Chen, Jiaxin and Chen, Guanqi and Li, Haofeng and Li, Guanbin and Chen, Fei},
  journal={Computers in biology and medicine},
  volume={155},
  pages={106389},
  year={2023},
  publisher={Elsevier}
}

@article{BreastLesion,
  title={Global guidance network for breast lesion segmentation in ultrasound images},
  author={Xue, Cheng and Zhu, Lei and Fu, Huazhu and Hu, Xiaowei and Li, Xiaomeng and Zhang, Hai and Heng, Pheng-Ann},
  journal={Medical image analysis},
  volume={70},
  pages={101989},
  year={2021},
  publisher={Elsevier}
}

@article{SegmentationChallenge,
  title={Ultrasound image segmentation: a survey},
  author={Noble, J Alison and Boukerroui, Djamal},
  journal={IEEE Transactions on medical imaging},
  volume={25},
  number={8},
  pages={987--1010},
  year={2006},
  publisher={IEEE}
}

@article{LayerSupervise,
  title={Ultrasound image segmentation: a deeply supervised network with attention to boundaries},
  author={Mishra, Deepak and Chaudhury, Santanu and Sarkar, Mukul and Soin, Arvinder Singh},
  journal={IEEE Transactions on Biomedical Engineering},
  volume={66},
  number={6},
  pages={1637--1648},
  year={2018},
  publisher={IEEE}
}

@article{Gobletnet,
  title={GobletNet: Wavelet-Based High-Frequency Fusion Network for Semantic Segmentation of Electron Microscopy Images},
  author={Zhou, Yanfeng and Li, Lingrui and Wang, Chenlong and Song, Le and Yang, Ge},
  journal={IEEE Transactions on Medical Imaging},
  year={2024},
  publisher={IEEE}
}

@article{CNN+Transformer2,
  title={Syn-Net: A Synchronous Frequency-perception Fusion Network for Breast tumor Segmentation in Ultrasound Images},
  author={Zhao, Guangzhe and Zhu, Xingguo and Wang, Xueping and Yan, Feihu and Guo, Maozu},
  journal={IEEE Journal of Biomedical and Health Informatics},
  year={2024},
  publisher={IEEE}
}

@inproceedings{ConvHigh1,
  title={High-frequency component helps explain the generalization of convolutional neural networks},
  author={Wang, Haohan and Wu, Xindi and Huang, Zeyi and Xing, Eric P},
  booktitle={Proceedings of the IEEE/CVF conference on computer vision and pattern recognition},
  pages={8684--8694},
  year={2020}
}

@article{DeeplearningUltraosound,
  title={Automated breast ultrasound lesions detection using convolutional neural networks},
  author={Yap, Moi Hoon and Pons, Gerard and Marti, Joan and Ganau, Sergi and Sentis, Melcior and Zwiggelaar, Reyer and Davison, Adrian K and Marti, Robert},
  journal={IEEE journal of biomedical and health informatics},
  volume={22},
  number={4},
  pages={1218--1226},
  year={2017},
  publisher={IEEE}
}

@article{AAU-Net,
  title={AAU-net: an adaptive attention U-net for breast lesions segmentation in ultrasound images},
  author={Chen, Gongping and Li, Lei and Dai, Yu and Zhang, Jianxun and Yap, Moi Hoon},
  journal={IEEE Transactions on Medical Imaging},
  volume={42},
  number={5},
  pages={1289--1300},
  year={2022},
  publisher={IEEE}
}

@article{EdgeBackgroundAware,
  title={SMU-Net: Saliency-guided morphology-aware U-Net for breast lesion segmentation in ultrasound image},
  author={Ning, Zhenyuan and Zhong, Shengzhou and Feng, Qianjin and Chen, Wufan and Zhang, Yu},
  journal={IEEE transactions on medical imaging},
  volume={41},
  number={2},
  pages={476--490},
  year={2021},
  publisher={IEEE}
}

@article{boundaryloss,
  title={LightCM-PNet: A lightweight pyramid network for real-time prostate segmentation in transrectal ultrasound},
  author={Wang, Weirong and Pan, Bo and Ai, Yue and Li, Gonghui and Fu, Yili and Liu, Yanjie},
  journal={Pattern Recognition},
  volume={156},
  pages={110776},
  year={2024},
  publisher={Elsevier}
}

@inproceedings{FoggyWavelet,
  title={Learning generalized segmentation for foggy-scenes by bi-directional wavelet guidance},
  author={Bi, Qi and You, Shaodi and Gevers, Theo},
  booktitle={Proceedings of the AAAI Conference on Artificial Intelligence},
  volume={38},
  number={2},
  pages={801--809},
  year={2024}
}

@article{WaveletDownsample,
  title={Haar wavelet downsampling: A simple but effective downsampling module for semantic segmentation},
  author={Xu, Guoping and Liao, Wentao and Zhang, Xuan and Li, Chang and He, Xinwei and Wu, Xinglong},
  journal={Pattern recognition},
  volume={143},
  pages={109819},
  year={2023},
  publisher={Elsevier}
}

@article{GaborWavelet,
  title={Gcsba-net: Gabor-based and cascade squeeze bi-attention network for gland segmentation},
  author={Wen, Zhijie and Feng, Ru and Liu, Jingxin and Li, Ying and Ying, Shihui},
  journal={IEEE Journal of Biomedical and Health Informatics},
  volume={25},
  number={4},
  pages={1185--1196},
  year={2020},
  publisher={IEEE}
}

@article{unet++,
  title={Unet++: Redesigning skip connections to exploit multiscale features in image segmentation},
  author={Zhou, Zongwei and Siddiquee, Md Mahfuzur Rahman and Tajbakhsh, Nima and Liang, Jianming},
  journal={IEEE transactions on medical imaging},
  volume={39},
  number={6},
  pages={1856--1867},
  year={2019},
  publisher={ieee}
}

@inproceedings{OneVOS,
  title={Onevos: unifying video object segmentation with all-in-one transformer framework},
  author={Li, Wanyun and Guo, Pinxue and Zhou, Xinyu and Hong, Lingyi and He, Yangji and Zheng, Xiangyu and Zhang, Wei and Zhang, Wenqiang},
  booktitle={European Conference on Computer Vision},
  pages={20--40},
  year={2024},
  organization={Springer}
}

@article{CAMUS,
  title={Deep learning for segmentation using an open large-scale dataset in 2D echocardiography},
  author={Leclerc, Sarah and Smistad, Erik and Pedrosa, Joao and {\O}stvik, Andreas and Cervenansky, Frederic and Espinosa, Florian and Espeland, Torvald and Berg, Erik Andreas Rye and Jodoin, Pierre-Marc and Grenier, Thomas and others},
  journal={IEEE transactions on medical imaging},
  volume={38},
  number={9},
  pages={2198--2210},
  year={2019},
  publisher={IEEE}
}

@inproceedings{tg3k,
  title={Comparison of thyroid segmentation techniques for 3D ultrasound},
  author={Wunderling, Tom and Golla, Bjoern and Poudel, Prabal and Arens, Christoph and Friebe, Michael and Hansen, Christian},
  booktitle={Medical Imaging 2017: Image Processing},
  volume={10133},
  pages={346--352},
  year={2017},
  organization={SPIE}
}

@article{aot,
  title={Associating objects with transformers for video object segmentation},
  author={Yang, Zongxin and Wei, Yunchao and Yang, Yi},
  journal={Advances in Neural Information Processing Systems},
  volume={34},
  pages={2491--2502},
  year={2021}
}

@article{zhang2025tcfnet,
  title={TCFNet: Bidirectional face-bone transformation via a Transformer-based coarse-to-fine point movement network},
  author={Zhang, Runshi and Jie, Bimeng and He, Yang and Wang, Junchen},
  journal={Medical Image Analysis},
  volume={105},
  pages={103653},
  year={2025},
  publisher={Elsevier}
}

@inproceedings{highfrequency2intra,
  title={Frequency-driven imperceptible adversarial attack on semantic similarity},
  author={Luo, Cheng and Lin, Qinliang and Xie, Weicheng and Wu, Bizhu and Xie, Jinheng and Shen, Linlin},
  booktitle={Proceedings of the IEEE/CVF conference on computer vision and pattern recognition},
  pages={15315--15324},
  year={2022}
}

@inproceedings{SENet,
  title={Squeeze-and-excitation networks},
  author={Hu, Jie and Shen, Li and Sun, Gang},
  booktitle={Proceedings of the IEEE conference on computer vision and pattern recognition},
  pages={7132--7141},
  year={2018}
}

@inproceedings{finegrainedaccuracy,
  title={Refinemask: Towards high-quality instance segmentation with fine-grained features},
  author={Zhang, Gang and Lu, Xin and Tan, Jingru and Li, Jianmin and Zhang, Zhaoxiang and Li, Quanquan and Hu, Xiaolin},
  booktitle={Proceedings of the IEEE/CVF conference on computer vision and pattern recognition},
  pages={6861--6869},
  year={2021}
}

@misc{mmseg2020,
    title={{MMSegmentation}: OpenMMLab Semantic Segmentation Toolbox and Benchmark},
    author={MMSegmentation Contributors},
    howpublished = {\url{https://github.com/open-mmlab/mmsegmentation}},
    year={2020}
}

@article{zhang2024utsrmorph,
  title={UTSRMorph: A Unified Transformer and Superresolution Network for Unsupervised Medical Image Registration},
  author={Zhang, Runshi and Mo, Hao and Wang, Junchen and Jie, Bimeng and He, Yang and Jin, Nenghao and Zhu, Liang},
  journal={IEEE Transactions on Medical Imaging},
  year={2024},
  publisher={IEEE}
}

@inproceedings{he2024ma,
  title={Ma-lmm: Memory-augmented large multimodal model for long-term video understanding},
  author={He, Bo and Li, Hengduo and Jang, Young Kyun and Jia, Menglin and Cao, Xuefei and Shah, Ashish and Shrivastava, Abhinav and Lim, Ser-Nam},
  booktitle={Proceedings of the IEEE/CVF Conference on Computer Vision and Pattern Recognition},
  pages={13504--13514},
  year={2024}
}

@inproceedings{zhou2024rmem,
  title={Rmem: Restricted memory banks improve video object segmentation},
  author={Zhou, Junbao and Pang, Ziqi and Wang, Yu-Xiong},
  booktitle={Proceedings of the IEEE/CVF Conference on Computer Vision and Pattern Recognition},
  pages={18602--18611},
  year={2024}
}

@article{yang2021associating,
  title={Associating objects with transformers for video object segmentation},
  author={Yang, Zongxin and Wei, Yunchao and Yang, Yi},
  journal={Advances in Neural Information Processing Systems},
  volume={34},
  pages={2491--2502},
  year={2021}
}

@article{chen2025frequency,
  title={Frequency-Dynamic Attention Modulation for Dense Prediction},
  author={Chen, Linwei and Gu, Lin and Fu, Ying},
  journal={arXiv preprint arXiv:2507.12006},
  year={2025}
}

@article{lin2024instrument,
  title={Instrument-tissue interaction detection framework for surgical video understanding},
  author={Lin, Wenjun and Hu, Yan and Fu, Huazhu and Yang, Mingming and Chng, Chin-Boon and Kawasaki, Ryo and Chui, Cheekong and Liu, Jiang},
  journal={IEEE Transactions on Medical Imaging},
  volume={43},
  number={8},
  pages={2803--2813},
  year={2024},
  publisher={IEEE}
}

@article{luo2025lgffm,
  title={LGFFM: A Localized and Globalized Frequency Fusion Model for Ultrasound Image Segmentation},
  author={Luo, Xiling and Wang, Yi and Ou-Yang, Le},
  journal={IEEE Transactions on Medical Imaging},
  year={2025},
  publisher={IEEE}
}

@article{deng2025echocardiography,
  title={Echocardiography video segmentation via neighborhood correlation mining},
  author={Deng, Xiaolong and Wu, Huisi},
  journal={IEEE Transactions on Medical Imaging},
  year={2025},
  publisher={IEEE}
}

@inproceedings{dandan2023semi,
  title={Semi supervised segmentation of thyroid based on ultrasound images with wavelet and boundaries features},
  author={Dandan, Li and Fei, Liu and Fangang, Meng and Yang, Du and Jing, Jin},
  booktitle={2023 IEEE International Instrumentation and Measurement Technology Conference (I2MTC)},
  pages={1--6},
  year={2023},
  organization={IEEE}
}

@article{xiong2025hcmnet,
  title={Hcmnet: A hybrid cnn-mamba network for breast ultrasound segmentation for consumer assisted diagnosis},
  author={Xiong, Youqiang and Shu, Xiu and Liu, Qiao and Yuan, Di},
  journal={IEEE Transactions on Consumer Electronics},
  year={2025},
  publisher={IEEE}
}

@inproceedings{zheng2024gwunet,
  title={GWUNet: A UNet with Gated Attention and Improved Wavelet Transform for Thyroid Nodules Segmentation},
  author={Zheng, Shuijing and Yu, Suxi and Wang, Yi and Wen, Jing},
  booktitle={International Conference on Multimedia Modeling},
  pages={31--44},
  year={2024},
  organization={Springer}
}

@article{liang2026wkpnet,
  title={WKPNet: A novel wavelet-KAN-POLA network for medical image segmentation},
  author={Liang, Pengchen and Zeng, Quanhong and Liang, Bocheng and Huang, Haishan and Zhang, Yudong and Pu, Bin and Chen, Jianguo},
  journal={Biomedical Signal Processing and Control},
  volume={113},
  pages={108988},
  year={2026},
  publisher={Elsevier}
}

@inproceedings{zhao2023ultrasound,
  title={Ultrasound video segmentation with adaptive temporal memory},
  author={Zhao, He and Men, Qianhui and Gleed, Alexander and Papageorghiou, Aris T and Noble, J Alison},
  booktitle={International Workshop on Advances in Simplifying Medical Ultrasound},
  pages={3--12},
  year={2023},
  organization={Springer}
}

@article{li2024cascaded,
  title={Cascaded inner-outer clip retformer for ultrasound video object segmentation},
  author={Li, Jialu and Zhu, Lei and Xing, Zhaohu and Zhao, Baoliang and Hu, Ying and Lv, Faqin and Wang, Qiong},
  journal={IEEE Journal of Biomedical and Health Informatics},
  year={2024},
  publisher={IEEE}
}

@inproceedings{tu2025spatial,
  title={Spatial-Temporal Memory Filtering SAM for Lesion Segmentation in Breast Ultrasound Videos},
  author={Tu, Zhengzheng and Zong, Liang and Jiang, Bo and Wang, Haowen and Wang, Kunpeng and Zhang, Chaoxue},
  booktitle={International Conference on Medical Image Computing and Computer-Assisted Intervention},
  pages={547--557},
  year={2025},
  organization={Springer}
}

@inproceedings{zeiler2014visualizing,
  title={Visualizing and understanding convolutional networks},
  author={Zeiler, Matthew D and Fergus, Rob},
  booktitle={European conference on computer vision},
  pages={818--833},
  year={2014},
  organization={Springer}
}

@inproceedings{li2025weaveseg,
  title={WeaveSeg: Iterative Contrast-weaving and Spectral Feature-refining for Nuclei Instance Segmentation},
  author={Li, Jiajia and Wu, Huisi and Qin, Jing},
  booktitle={Proceedings of the IEEE/CVF International Conference on Computer Vision},
  pages={21984--21993},
  year={2025}
}





\end{document}